\documentclass[10pt,journal,compsoc]{IEEEtran}
\usepackage{xspace}
\usepackage[dvipsnames]{xcolor}
\usepackage[bookmarks=false]{hyperref}

\newcommand\myshade{85}
\colorlet{mylinkcolor}{violet}
\colorlet{mycitecolor}{YellowOrange}
\colorlet{myurlcolor}{Aquamarine}
\hypersetup{
  linkcolor  = mylinkcolor!\myshade!black,
  citecolor  = mycitecolor!\myshade!black,
  urlcolor   = myurlcolor!\myshade!black,
  colorlinks = true,
}

\usepackage{graphicx}

\graphicspath{{gfx/},{plots/}}
\DeclareGraphicsExtensions{.pdf,.jpeg,.png}
\providecommand{\norm}[1]{\lVert#1\rVert}

\ifCLASSOPTIONcompsoc
  \usepackage[nocompress]{cite}
\else
  \usepackage{cite}
\fi

\usepackage[cmex10]{amsmath}
\usepackage{amsfonts}
\usepackage{multirow}
\usepackage{rotating}
\usepackage{booktabs}

\usepackage{algorithmic}
\usepackage{algorithm}
\usepackage{array}
\usepackage{mdwmath}
\usepackage{mdwtab}
\usepackage{calc}
\ifCLASSOPTIONcompsoc
 \usepackage[caption=false,font=normalsize,labelfont=sf,textfont=sf]{subfig}
\else
 \usepackage[caption=false,font=footnotesize]{subfig}
\fi

\usepackage{url}
\usepackage{tikz}

\DeclareMathOperator{\argmin}{argmin}

\def\cmupie{CMU MultiPIE~}

\definecolor{emotion}{RGB}{255, 138, 101}
\definecolor{pose}{RGB}{255, 183, 77}
\definecolor{identity}{RGB}{33, 129, 211}

\DeclareRobustCommand{\tikzcircle}[1]{\tikz{\filldraw[draw=#1,fill=#1] (0,0) circle [radius=0.1cm];}}

\DeclareRobustCommand{\tikzplus}[1]{{\color{#1}\bfseries +}}

\def\mawsf{WSF-MA\xspace}
\def\wsf{WSF\xspace}
\newcommand{\dwsf}{{Deep WSF}\xspace}
\newcommand{\tr}{\xspace\text{Tr}}

\def\Cdeep{\mathcal{C}_{\text{deep}}}
\def\Cwdeep{\mathcal{C}_{\text{dwsf}}}
\def\etal{{\it et al.}}

\begin{document}

\title{A deep matrix factorization method for learning attribute representations}

% use \thanks{} to gain access to the first footnote area
% a separate \thanks must be used for each paragraph as LaTeX2e's \thanks
% was not built to handle multiple paragraphs
%
%
%\IEEEcompsocitemizethanks is a special \thanks that produces the bulleted
% lists the Computer Society journals use for "first footnote" author
% affiliations. Use \IEEEcompsocthanksitem which works much like \item
% for each affiliation group. When not in compsoc mode,
% \IEEEcompsocitemizethanks becomes like \thanks and
% \IEEEcompsocthanksitem becomes a line break with idention. This
% facilitates dual compilation, although admittedly the differences in the
% desired content of \author between the different types of papers makes a
% one-size-fits-all approach a daunting prospect. For instance, compsoc 
% journal papers have the author affiliations above the "Manuscript
% received ..."  text while in non-compsoc journals this is reversed. Sigh.

\author{George~Trigeorgis, \and
        Konstantinos~Bousmalis,~\IEEEmembership{Student Member,~IEEE,} \and 
        Stefanos~Zafeiriou,~\IEEEmembership{Member,~IEEE} \and
        Bj\"orn~W.~Schuller,~\IEEEmembership{Senior member,~IEEE}%
  \IEEEcompsocitemizethanks{
    \IEEEcompsocthanksitem 
      G. Trigeorgis, S. Zafeiriou, and B. W. Schuller are with the Department of Computing,
        Imperial College London, SW7 2RH, London, UK\protect\\
      E-mail: \href{mailto:{g.trigeorgis,z.zafeiriou,bjoern.schuller}@imperial.ac.uk}{g.trigeorgis@imperial.ac.uk}
    \IEEEcompsocthanksitem 
      K. Bousmalis is with Google Robotics\protect\\
      E-mail: \href{mailto:konstantinos@google.com}{konstantinos@google.com}
    }
}

\newcommand{\deepseminmf}{Deep \seminmf\xspace} 
\newcommand{\multipie}{Multi-PIE\xspace}
\newcommand{\xmvts}{XM2VTS\xspace}
\newcommand{\ZsT}{{\boldsymbol Z}_m^{\top} {\boldsymbol Z}_{m-1}^{\top} \cdots {\boldsymbol Z}_1^{\top}}
\newcommand{\Ta}{*}
\newcommand{\Zt}[1]{\tilde{Z}_{#1}^\Ta}
\newcommand{\boldX}{\boldsymbol X}
\newcommand{\boldZ}{\boldsymbol Z}
\newcommand{\boldH}{\boldsymbol H}
\newcommand{\boldN}{\boldsymbol N}
\newcommand{\boldR}{\boldsymbol R}
\newcommand{\seminmf}{Semi-NMF\xspace}
\newcommand{\citet}[1]{\cite{#1}}
\newcommand{\bs}[1]{\boldsymbol{#1}}

% for Computer Society papers, we must declare the abstract and index terms
% PRIOR to the title within the \IEEEcompsoctitleabstractindextext IEEEtran
% command as these need to go into the title area created by \maketitle.
\IEEEcompsoctitleabstractindextext{%
\begin{abstract}
Semi-Non-negative Matrix Factorization is a technique that learns a low-dimensional representation of a dataset that lends itself to a clustering interpretation. It is possible that the mapping between this new representation and our original data matrix contains rather complex hierarchical information with implicit lower-level hidden attributes, that classical one level clustering methodologies can not interpret. In this work we propose a novel model, Deep Semi-NMF, that is able to learn such hidden representations that allow themselves to an interpretation of clustering  according to different, unknown attributes of a given dataset. We also present a semi-supervised version of the algorithm, named Deep WSF, that allows the use of (partial) prior information for each of the known attributes of a dataset, that allows the model to be used on datasets with mixed attribute knowledge. Finally, we show that our models are able to learn low-dimensional representations that are better suited for clustering, but also classification, outperforming Semi-Non-negative Matrix Factorization, but also other state-of-the-art methodologies variants. 
\end{abstract}

}

% make the title area
\maketitle

% To allow for easy dual compilation without having to reenter the
% abstract/keywords data, the \IEEEcompsoctitleabstractindextext text will
% not be used in maketitle, but will appear (i.e., to be "transported")
% here as \IEEEdisplaynotcompsoctitleabstractindextext when compsoc mode
% is not selected <OR> if conference mode is selected - because compsoc
% conference papers position the abstract like regular (non-compsoc)
% papers do!
\IEEEdisplaynotcompsoctitleabstractindextext
% \IEEEdisplaynotcompsoctitleabstractindextext has no effect when using
% compsoc under a non-conference mode.

% For peer review papers, you can put extra information on the cover
% page as needed:
% \ifCLASSOPTIONpeerreview
% \begin{center} \bfseries EDICS Category: 3-BBND \end{center}
% \fi
%
% For peerreview papers, this IEEEtran command inserts a page break and
% creates the second title. It will be ignored for other modes.
\IEEEpeerreviewmaketitle

\section{Introduction}
\IEEEPARstart{M}{atrix} factorization is a particularly useful family of techniques in data analysis. In recent years, there has been a significant amount of research on factorization methods that focus on particular characteristics of both the data matrix and the resulting factors. Non-negative matrix factorization (NMF), for example, focuses on the decomposition of non-negative multivariate data matrix $\boldX$ into factors $\boldZ$ and $\boldH$ that are also non-negative, such that $\boldX \approx \boldZ \boldH$. The application area of the family of NMF algorithms has grown significantly during the past years. It has been shown that they can be a successful dimensionality reduction technique over a variety of areas including, but not limited to, environmetrics \cite{paatero1994positive}, microarray  data analysis \cite{brunet2004metagenes,devarajan2008nonnegative}, document clustering \cite{berry2005email}, face recognition \cite{zafeiriou2006exploiting,kotsia2007novel}, blind audio source separation \cite{weninger2012optimization} and more. What makes NMF algorithms particularly attractive is the non-negativity constraints imposed on the factors they produce, allowing for better interpretability. Moreover, it has been shown that NMF variants (such as the \seminmf) are equivalent to a soft version of \emph{k}-means clustering, and that in fact, NMF variants are expected to perform better than \emph{k}-means clustering particularly when the data is not distributed in a spherical manner~\cite{ding2010convex,ding2005equivalency}. 

In order to extend the applicability of NMF in cases where our data matrix $\boldX$ is not strictly non-negative, \cite{ding2010convex} introduced \seminmf, an NMF variant that imposes non-negativity constraints only on the second factor $\boldH$, but allows mixed signs in both the data matrix $\boldX$ and the first factor $\boldZ$. This was motivated from a clustering perspective, where $\boldZ$ represents cluster centroids, and $\boldH$ represents soft membership indicators for every data point, allowing \seminmf to learn new lower-dimensional features from the data that have a convenient clustering interpretation.

\begin{figure}[tb]
\subfloat[\seminmf]{\includegraphics[width=0.4\linewidth]{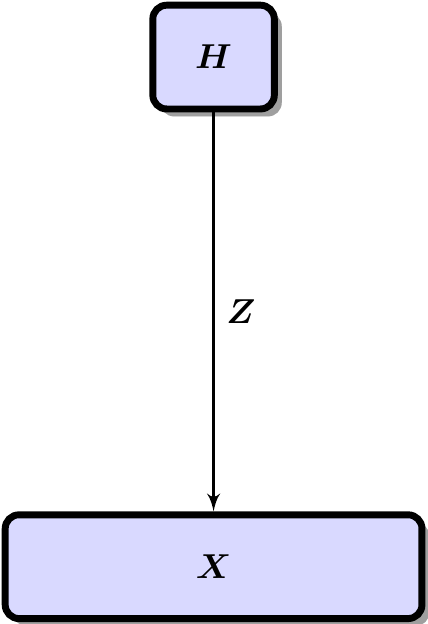}\label{fig:simplenmf}}
\hspace{1cm}
\subfloat[Deep \seminmf]{\includegraphics[width=0.4\linewidth]{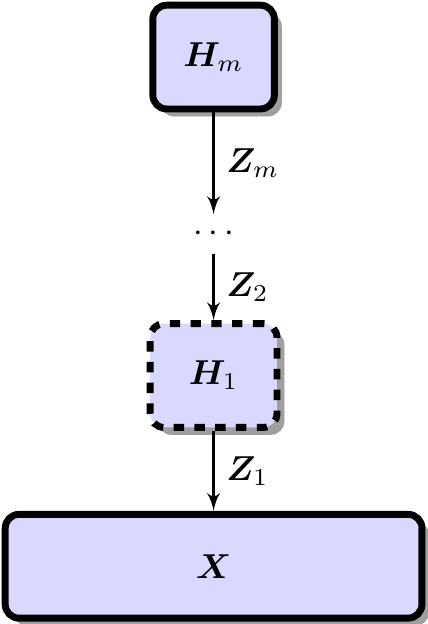}\label{fig:deepnmf}}
\caption{\protect\subref{fig:simplenmf} A \seminmf model results in a linear transformation of the initial input space. \protect\subref{fig:deepnmf} Deep Semi-NMF learns a hierarchy of hidden representations that aid in uncovering the final lower-dimensional representation of the data.}
\label{fig:deepexample}
\end{figure}

It is possible that the mapping $\boldZ$ between this new representation $\boldH$ and our original data matrix $\boldX$ contains rather complex hierarchical and structural information. Such a complex dataset $\boldX$ is produced by a multi-modal data distribution which is a mixture of several distributions, where each of these constitutes an {\it attribute} of the dataset. Consider for example the problem of mapping images of faces to their identities: a face image also contains information about attributes like pose and expression that can help identify the person depicted. One could argue that by further factorizing this mapping $\boldZ$, in a way that each factor adds an extra layer of abstraction, one could automatically learn such latent attributes and the intermediate hidden representations that are implied, allowing for a better higher-level feature representation $\boldH$. In this work, we propose Deep Semi-NMF, a novel approach that is able to factorize a matrix into multiple factors in an unsupervised fashion -- see \autoref{fig:deepexample}, and it is therefore able to learn multiple hidden representations of the original data. 
As \seminmf has a close relation to \emph{k}-means clustering, Deep \seminmf also has a clustering interpretation according to the different latent attributes of our dataset, as demonstrated in \autoref{fig:intuition}. 
Using a non-linear deep model for matrix factorization also allows us to project data-points which are not initially linearly separable into a representation that is; fact which we demonstrate in \autoref{sub:xor_problem}.

It might be the case that the different attributes of our data are not latent. If those are known and we actually have some label information about some or all of our data, we would naturally want to leverage it and learn representations that would make the data more separable according to each of these attributes. To this effect, we also propose a weakly-supervised Deep \seminmf (\dwsf), a technique that is able to learn, in a semi-supervised manner, a hierarchy of representations for a given dataset. Each level of this hierarchy corresponds to a specific attribute that is known a priori, and we show that by incorporating partial label information via graph regularization techniques we are able to perform better than with a fully unsupervised \deepseminmf in the task of classifying our dataset of faces according to different attributes, when those are known.  We also show that by initializing an unsupervised \deepseminmf with the weights learned by a \dwsf we are able to improve the clustering performance of the \deepseminmf. This could be particularly useful if we have, as in our example, a small dataset of images of faces with partial attribute labels and a larger one with no attribute labels. By initializing a \deepseminmf with the weights learned with \dwsf from the small labelled dataset we can leverage all the information we have and allow our unsupervised model to uncover better representations for our initial data on the task of clustering faces.

Relevant to our proposal are hierarchical clustering algorithms \cite{Herrero2001,Zhao2005} which are popular in gene and document clustering applications.
These algorithms typically abstract the initial data distribution as a form of tree called a {\it dendrogram}, which is useful for analysing the data and help identify genes that can be used as biomarkers or topics of a collection of documents. This makes it hard to incorporate out-of-sample data and prohibits the use of other techniques than clustering. 

Another line of work which is related to ours is multi-label learning \cite{Tsoumakas2007}. Multi-label learning techniques rely on the correlations \cite{Zhang2010} that exist between different attributes to extract better features. We are not interested in cases where there is complete knowledge about each of the attributes of the dataset but rather we propose a new paradigm of learning representations where have data with only partly annotated attributes. An example of this is a mixture of datasets where each one has label information about a different set of attributes. In this new paradigm we can {\it not} leverage the correlations between the attribute labels and we rather rely on the hierarchical structure of the data to uncover relations between the different dataset attributes.  To the best of our knowledge this is the first piece of work that tries to automatically discover the representations for different (known and unknown) attributes of a dataset with an application to a multi-modal application such as face clustering.

The novelty of this work can be summarised as follows: \emph{(1)} we outline a novel deep framework \footnote{A preliminary version of this work has appeared in \citet{Trigeorgis2014}.} for matrix factorization suitable for clustering of multimodally distributed objects such as faces, \emph{(2)} we present a greedy  algorithm to optimize the factors of the \seminmf problem, inspired by recent advances in deep learning~\cite{hinton2006reducing},
\emph{(3)} we evaluate the representations learned by different NMF--variants in terms of clustering performance, \emph{(4)} present the \dwsf{} model that can use already known (partial) information for the attributes of our data distribution to extract better features for our model, and \emph{(5)} demonstrate how to improve the performance of \deepseminmf, by using the existing weights from a trained \dwsf{} model.

\section{Background} \label{sec:related}
\def\algorithmautorefname{Algorithm}
In this work, we assume that our data is provided in a matrix form ${\boldX}\in \mathbb{R}^{p\times n}$, 
i.e., ${\bs X}=[{\bs x_1}, {\bs x_2}, \ldots, {\bs x_n}]$ is a collection of $n$ data vectors as columns, each with $p$ features. Matrix factorization aims at finding factors of ${\bs X}$ that satisfy certain constraints. In Singular Value Decomposition (SVD)~\cite{golub1970singular}, the method that underlies Principal Component Analysis (PCA)~\cite{wold1987principal}, we factorize ${\boldX}$ into two factors: the loadings or bases $\boldZ \in \mathbb{R}^{p\times k}$ and the features or components $\boldH \in {\mathbb{R}}^{k\times n}$, without imposing any sign restrictions on either our data or the resulting factors. In Non-negative Matrix Factorization (NMF)~\cite{seung2001algorithms} we assume that all matrices involved contain only non-negative elements\footnote{When not clear from the context we will use the notation $\bs A^+$ to state that a matrix ${\bs A}$ contains only non-negative elements. Similarly, when not clear, we will use the notation $\bs A^\pm$ to state that ${\bs A}$ may contain any real number.}, so we try to approximate a factorization ${\bs X^+} \approx {\boldZ^+ \boldH^+}$.
\subsection{\seminmf}

% \begin{algorithm}[tb]
%     \caption{\textbf{The \seminmf algorithm:} given initial data matrix ${\bs X} \in \mathbb{R}^{p\times n}$ and the number of components $k$, \seminmf decomposes it into two matrices ${\bs Z} \in \mathbb{R}^{p\times k}$ and ${\bs H} \in {\mathbb{R}^+_0}^{k\times n}$ }

%       \begin{algorithmic}  
%         % \Function {\textsc{SemiNMF} }
%             \STATE {\bfseries Input:} {${\bs X} \in \mathbb{R}^{p\times n}$, $k$}
%             \STATE {\bfseries Output:} {${\bs Z} \in \mathbb{R}^{p\times k}$,${\bs H} \in {\mathbb{R}^+_0}^{k\times n}$}
%             \STATE Initialise ${\bs H}$
%             \REPEAT 
%                 \STATE{${\boldZ} \gets {\boldX  \boldH}^{\dagger}$}
%                 \STATE{${\boldH} \gets {\boldH} \odot \left[{\dfrac{[{\boldZ}^\top {\bs X}]^{\text{pos}} + {[{\boldZ}^\top {\bs Z}]}^{\text{neg}}  {\bs H}}{[{\boldZ}^\top {\bs X}]^{\text{neg}} + {[{\boldZ}^\top {\bs Z}]}^{\text{pos}} {\bs H} + \epsilon}}\right]^{\eta}$}
%             \UNTIL{Stopping criterion is reached}
%         % \ENDFUNCTION
%       \end{algorithmic}   
%       \label{alg:seminmf_algo}
% \end{algorithm}

In turn, \seminmf~\cite{ding2010convex}  relaxes the non-negativity constrains of NMF and allows the data matrix ${\boldX}$ and the loadings matrix ${\boldZ}$ to have mixed signs, while restricting only the features matrix ${\boldH}$ to comprise of strictly non-negative components, thus approximating the following factorization: 
\begin{equation}
{\boldX}^{\pm} \approx {\boldZ^{\pm} \boldH^+}.
\label{eq:seminmf}
\end{equation}
This is motivated  from a clustering perspective. If we view ${{\bs Z} = [{\bs z_1}, {\bs z_2}, \ldots, {\bs z_k}]}$ as the cluster centroids, then ${{\bs H} = [{\bs h_1}, {\bs h_2}, \ldots, {\bs h_n}]}$ can be viewed as the cluster indicators for each datapoint. 

In fact, if we had a matrix ${\boldH}$ that was not only non-negative but also orthogonal, such that ${{\boldH \boldH^T}={\bs I}}$ \cite{ding2010convex}, then every column vector would have only one positive element, making \seminmf equivalent to $k$-means, with the following cost function:
\begin{equation}
C_{k-\text{means}} = \sum_{i=1}^n \sum_{j=1}^k h_{ki} \|{\bs x_i} - {\bs z_k}\|^2 = \| {\bs X} - {\boldZ \boldH} \|^2_F
\end{equation}
where $\|\cdot\|$ denotes the $L_2$-norm of a vector and $\|\cdot\|_F$ the Frobenius norm of a matrix.

Thus \seminmf, which does not impose an orthogonality constraint on its features matrix, can be seen as a soft clustering method where the features matrix describes the compatibility of each component with a cluster centroid, a base in $\bs Z$. In fact, the cost function we optimize for approximating the \seminmf factors is indeed:
\begin{equation}
C_{\text{\seminmf}}=\|{\bs X} - {\bs Z}{\bs H}\|^2_F .
\end{equation}
We optimize $C_{\text{\seminmf}}$ via an alternate optimization of $\bs Z^\pm$ and $\bs H^+$: we iteratively update each of the factors while fixing the other, imposing the non-negativity constrains only on the features matrix $\boldH$:
\begin{equation}
{\boldZ} \gets {\boldX \boldH}^{\dagger}
\end{equation}
where ${\bs H}^{\dagger}$ is the Moore--Penrose pseudo-inverse of ${\bs H}$, and
\begin{IEEEeqnarray}{r}
    {\boldH} \gets {\boldH} \odot \sqrt{\dfrac{[{\boldZ}^\top {\bs X}]^{\text{pos}} + {[{\boldZ}^\top {\bs Z}]}^{\text{neg}}  {\bs H}}{{[{\boldZ}^\top {\bs X}]}^{\text{neg}} + {[{\boldZ}^\top {\bs Z}]}^{\text{pos}} {\bs H}}}
\end{IEEEeqnarray}
where $\epsilon$ is a small number to avoid division by zero, ${\bs A}^{\text{pos}}$ is a matrix that has the negative elements of matrix $\bs A$ replaced with 0, and similarly ${\bs A}^{\text{neg}}$ is one that has the positive elements of ${\bs A}$ replaced with 0:
\begin{equation}
  \forall i,j\;.\;\mathbf{A}^{\text{pos}}_{ij} = \dfrac{|\mathbf{A}_{ij}| + \mathbf{A}_{ij}}{2}, \quad
  \mathbf{A}^{\text{neg}}_{ij} = \dfrac{|\mathbf{A}_{ij}| - \mathbf{A}_{ij}}{2}.
\end{equation}

\subsection{State-of-the-art for learning features for clustering based on NMF-variants}
In this work, we compare our method with, among others, the state-of-the-art NMF techniques for learning features for the purpose of clustering. \citet{cai2011graph} proposed a graph-regularized NMF
(GNMF) which takes into account the intrinsic geometric and discriminating structure of the data space, which is essential to the real-world applications, especially in the area of clustering. To accomplish this, GNMF constructs a nearest neighbor graph to model the manifold structure. By preserving the graph structure, it allows the learned features to have more discriminating power than the standard NMF algorithm, in cases that the data are sampled from a submanifold which lies in a higher dimensional ambient space.

Closest to our proposal is recent work that has presented NMF-variants that factorize $\boldX$ into more than 2 factors. Specifically,~\citet{ahn2004multiplicative} have demonstrated the concept of Multi-Layer NMF on a set of facial images and~\cite{lyu2013algorithms,Cichocki06multilayernonnegative,song2013hierarchical} have proposed similar NMF models that can be used for Blind Source Separation, classification of digit images (MNIST), and documents. The representations of the Multi-layer NMF however do not lend themselves to a clustering interpretation, as the representations learned from our model.
Although the Multi-layer NMF is a promising technique for learning hierarchies of features from data, we show in this work that our proposed model, the Deep \seminmf outperforms the Multi-layer NMF and, in fact, all models we compared it with on the task of feature learning for clustering images of faces. 

\subsection{Semi-supervised matrix factorization}

For the case of the proposed \dwsf{} algorithms, we also evaluate our method with previous semi-supervised non-negative matrix factorization techniques. These include the Constrained Nonnegative Matrix Factorization ({CNMF}) \cite{Liu2012}, and the Discriminant Nonnegative Matrix Factorization (DNMF) \cite{Kotsia2007}. Although both take label information as additional constraints, the difference between these is that CNMF uses the label information as hard constrains on the resulting features $\bs H$, whereas DNMF tries to use the Fisher Criterion in order to incorporate discriminant information  in the decomposition \cite{Kotsia2007}. Both approaches only work for cases where we want to encode the prior information of only one attribute, in contrast to the proposed \dwsf{} model.

\section{Deep \seminmf}\label{sec:deepseminmf}

\begin{figure*}
  \begin{center}
      \includegraphics[width=\linewidth]{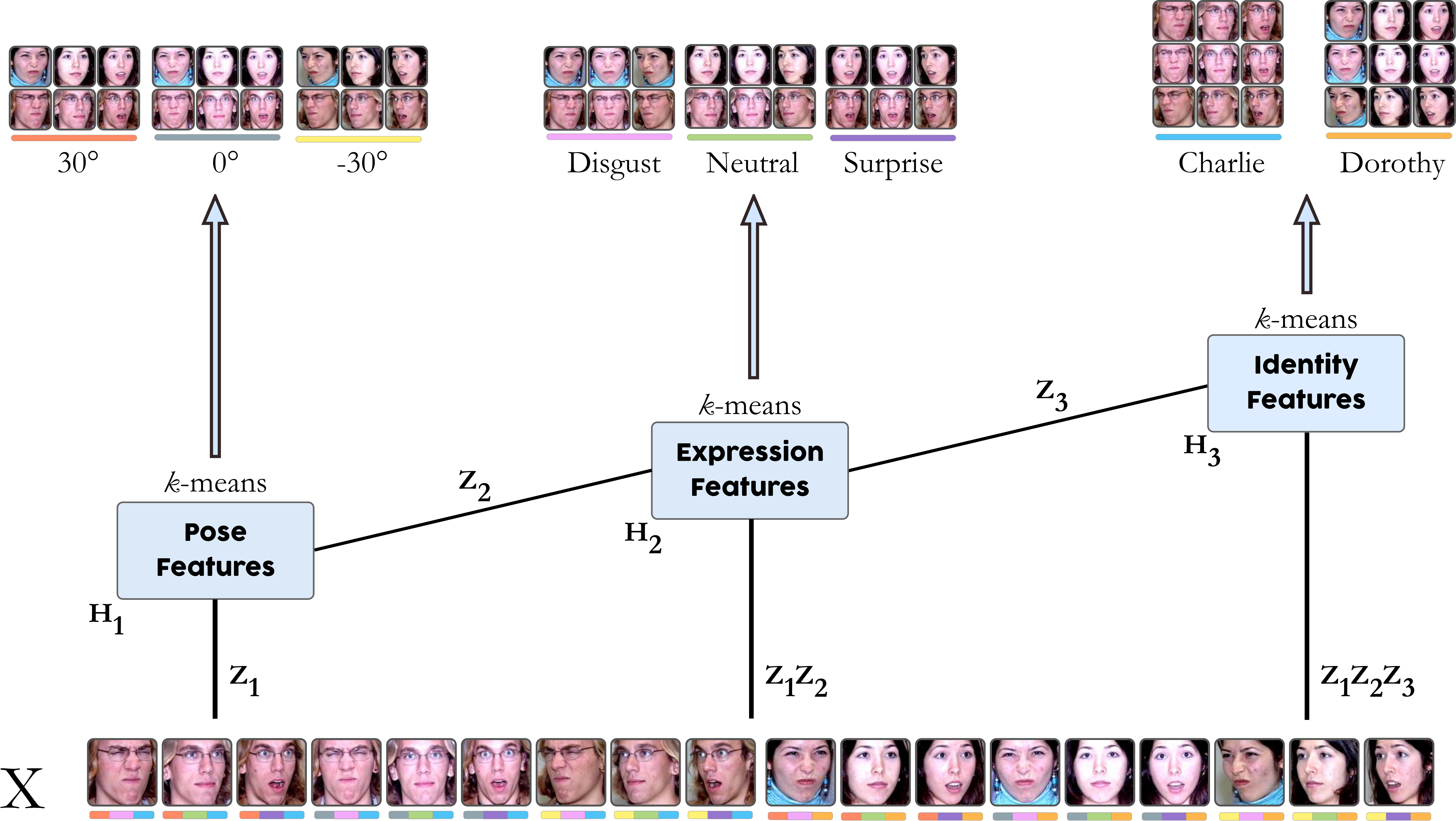}      
  \end{center}
  \caption{A Deep \seminmf model learns a hierarchical structure of features, with each layer learning a representation suitable for clustering according to the different attributes of our data. In this simplified, for demonstration purposes, example from the CMU Multi-PIE database, a Deep \seminmf model is able to simultaneously learn features for pose clustering ($\boldH_1$), for expression clustering ($\boldH_2$), and for identity clustering ($\boldH_3$). Each of the images in {\bf X} has an associated colour coding that indicates its memberships according to each of these attributes (pose/expression/identity).}
  \label{fig:intuition}
\end{figure*}

In \seminmf the goal is to construct a low-dimensional representation $\bs H^+$ of our original data $\bs X^\pm$, with the bases matrix $\bs Z^\pm$ serving as the mapping
between our original data and its lower-dimensional representation (see \autoref{eq:seminmf}). In many cases the data we wish to analyze is often rather complex and has a collection of distinct, often unknown, attributes. In this work for example, we deal with datasets of human faces where the variability in the data does not only stem from the difference in the appearance of the subjects, but also from other attributes, such as the pose of the head in relation to the camera, or the facial expression of the subject. The multi-attribute nature of our data calls for a hierarchical framework that is better at representing it than a shallow \seminmf.

We therefore propose here the Deep \seminmf model, which factorizes a given data matrix $\bs X$ into $m+1$ factors, as follows:
\begin{equation}
{\bs X}^{\pm} \approx {\bs Z}_1^{\pm}{\bs Z}_2^{\pm}\cdots{\bs Z}_m^{\pm}{\bs H}^+_m
\end{equation}
This formulation, as shown directly in \autoref{eq:expansion} with respect to Figures \ref{fig:deepexample} and \ref{fig:intuition} allows for a hierarchy of $m$ layers of implicit representations of our data that can be given by the following factorizations:

\begin{IEEEeqnarray}{rl}\nonumber
{\boldH}_{m-1}^+ \, &\approx {\boldZ}_m^{\pm} {\boldH}_m^{+}\\\nonumber
& \mathrel{\makebox[\widthof{=}]{\vdots}}  \\\nonumber
{\boldH}_2^+ &\approx {\bs Z}_3^{\pm} \cdots {\bs Z}_m^{\pm} {\bs H}_m^+ \\
{\boldH}_1^+ &\approx {\bs Z}_2^{\pm} \cdots {\bs Z}_m^{\pm} {\bs H}_m^+
\end{IEEEeqnarray}
As one can see above, we further restrict these implicit representations ($\boldH_1^+, \ldots, \boldH_{m-1}^+$) to also be non-negative. By doing so, every layer of this hierarchy of representations
also lends itself to a clustering interpretation, which constitutes our method radically different to other multi-layer NMF approaches \cite{lyu2013algorithms,Cichocki06multilayernonnegative,song2013hierarchical}. By examining \autoref{fig:intuition}, one can better understand the intuition of how that happens. In this case the input to the model, $\bs
X$, is a collection of face images from different subjects (identity), expressing a variety of facial expressions taken from many angles (pose). A \seminmf model would find a representation $\bs H$ of $\bs X$, which would be useful for performing clustering according to the identity of the subjects, and $\bs Z$ the mapping between these identities and the face images. A Deep \seminmf model also finds a representation of our data that has a similar interpretation at the top layer, its last factor $\boldH_m$. However, the mapping from identities to face images is now further analyzed as a product of three factors ${\boldZ = {\boldZ_1 \boldZ_2 \boldZ_3}}$, with ${\bs Z_3}$ corresponding to the mapping of identities to expressions, $\bs Z_2 \boldZ_3$ corresponding to the mapping of identities to poses, and finally $\boldZ_1 \boldZ_2 \boldZ_3$ corresponding to the mapping of identities to the face images. That means that, as shown in \autoref{fig:intuition} we are able to decompose our data in 3 different ways according to our 3 different attributes:
\begin{align}\nonumber
{\bs X}^{\pm} &\approx {\bs Z}_1^{\pm} {\bs H}_1^{+}\\\nonumber
{\bs X}^{\pm} &\approx {\bs Z}_1^{\pm} {\bs Z}_2^{\pm} {\bs H}_2^{+}\\
{\bs X}^{\pm} &\approx {\bs Z}_1^{\pm} {\bs Z}_2^{\pm} {\bs Z}_3^{\pm} {\bs H}_3^{+} \label{eq:expansion}
\end{align}
More over, due to the non-negativity constrains we enforce on the latent features $\bs H_{(\cdot)}$, it should be noted that this model does not collapse to a \seminmf model. Our hypothesis is that by further factorizing $\bs Z$ we are able to construct a deep model that is able to \textit{(1)} automatically learn what this latent hierarchy of attributes is; \textit{(2)} find representations of the data that are most suitable for clustering according to the attribute that corresponds to each layer in the model; and \textit{(3)} find a better high-level, final-layer representation for clustering according to the attribute with the lowest variability, in our case the identity of the face depicted. In our example in \autoref{fig:intuition} we would expect to find better features for clustering according to identities ${\bs H}_3^+$ by learning the hidden representations at each layer most suitable for each of the attributes in our data, in this example: ${\bs H}_1^+\approx{\bs Z}_2^\pm{\bs Z}_3^\pm{\bs H}_3^+$ for clustering  our original images in terms of poses and ${\bs H}_2^+\approx{\bs Z}_3^\pm{\bs H}_3^+$ for clustering the face images in terms of expressions.

\newcommand{\Zs}{{\bs Z}_1 {\bs Z}_2 \cdots {\bs Z}_m}

In order to expedite the approximation of the factors in our model, we pretrain each of the layers to have an initial approximation of the matrices ${\bs Z}_i, {\bs H}_i$  as this greatly improves the training time of the model. This is a tactic that has been employed successfully before \cite{hinton2006reducing} on deep autoencoder networks. To perform the pre-training, we first decompose the initial data matrix ${\boldX \approx {\bs Z}_1 {\bs H}_1}$, where ${\bs Z}_1 \in \mathbb{R}^{p\times k_1}$ and ${\bs H}_1 \in {\mathbb{R}^+_0}^{k_1\times n}$. Following this, we decompose the features matrix ${\boldH_1 \approx {\bs Z}_2 \boldH_2}$, where ${\bs Z}_2 \in \mathbb{R}^{k_1\times k_2}$ and ${\boldH_1 \in {\mathbb{R}^+_0}^{k_2\times n}}$, continuing to do so until we have pre-trained all of the layers. Afterwards, we can fine-tune the weights of each layer, by employing alternating minimization (with respect to the objective function in \autoref{eq:deep_nmf_linear_cost}) of the two factors in each layer, in order to reduce the total reconstruction error of the model, according to the cost function in \autoref{eq:deep_nmf_linear_cost}.
\begin{align}
\notag C_{\text{deep}} &= \dfrac{1}{2} \norm{\boldX - \Zs \boldH_m}^2_F\\ 
\notag &= tr[\boldX^\top \boldX - 2 \boldX^\top \Zs \boldH_m \\ 
& \quad + \boldH_m^{\top} \ZsT \Zs \boldH_m]
\label{eq:deep_nmf_linear_cost}
\end{align}

\begin{algorithm}[tb]
\caption{Suggested algorithm for training a Deep \seminmf model\protect\footnotemark. Initially we approximate the factors greedily using the \textsc{\seminmf} algorithm \cite{ding2010convex} and we fine-tune the factors until we reach the convergence criterion.}

    \begin{algorithmic}
    \label{algo:deepnmfalgo}
    % \FUNCTION {\textsc{DeepSemiNMF}}
        \STATE {\bfseries Input:} {${\bs X} \in \mathbb{R}^{p\times n}$, list of layer sizes }
        \STATE {\bfseries Output:} {weight matrices ${\bs Z}_i$ and feature matrices ${\bs H}_i$ for each of the layers}
        \STATE {}
        \STATE {Initialize Layers}
        \FORALL{layers}
            \STATE ${\bs Z}_i, {\bs H}_i \gets  $ {\sc SemiNmf}(${\bs H}_{i-1}$, layers($i$)) 
        \ENDFOR

        \STATE{}

        \REPEAT
            \FORALL{layers}
                \STATE
       $\tilde{{\bs H}_i} \gets \left\{
          \begin{array}{l l}
            {\bs H}_i           & \quad \text{\textbf{if}} \quad i = k \\
            {\bs Z}_{i+1} \tilde{\bs H}_{i+1} & \quad \text{otherwise}\\
          \end{array}\right.$ \\
       ${\bs \Psi} \gets  \prod_{k=1}^{i-1}{{\bs Z}_k}$ \\
       ${\bs Z}_i \gets {\bs \Psi}^{\dagger} {\bs X} \tilde{\bs H}_i^{\dagger}$ \\
        ${\bs H}_i \gets  {\bs H}_i \odot \left[ \dfrac{[{\bs \Psi}^{\top} {\bs X}]^{\text{pos}} + [{\bs \Psi}^{\top} {\bs \Psi}]^{\text{neg}} {\bs H}_i }{[{\bs \Psi}^{\top} {\bs X}]^{\text{neg}} + [{\bs \Psi}^{\top} {\bs \Psi}]^{\text{pos}} {\bs H}_i }\right]^{\eta}$
            \ENDFOR
        \UNTIL{Stopping criterion is reached}
    % \ENDFUNCTION
    \end{algorithmic}
\end{algorithm}

\textbf{Update rule for the weights matrix $\bs{Z}$}
We fix the rest of the weights for the $i_{th}$ layer and we minimize the cost function with respect to $\boldZ_i$. That is, we set $\dfrac{\partial C_\text{deep}}{{\partial {\boldZ}_i}} = 0$, 
which gives us the updates:
\begin{align}
\notag {\bs Z}_i &=  (\bs\Psi^{\top} \bs\Psi)^{-1} \bs\Psi^{\top} \boldX \tilde{\boldH_i}^{\top} (\tilde{\boldH_i} \tilde{\boldH_i}^{\top})^{-1}  \\
{\bs Z}_i &= \bs\Psi^{\dagger} \boldX \tilde{\boldH}^{\dagger}_i
\label{eq:Z_update_rule}
\end{align}

where $\bs\Psi = {\bs Z}_1 \cdots {\bs Z}_{i-1}$, $\dagger$ denotes the Moore-Penrose pseudo-inverse and $\tilde{\boldH}_i$ is the reconstruction of the $i^{th}$ layer's feature matrix.

\textbf{Update rule for features matrix $\boldH$} Utilizing a similar proof to \cite{ding2010convex}, {} we can formulate the update rule for $\boldH_i$ which enforces the non-negativity of $\bs H_i$: 
\begin{equation}
  \boldH_i = \boldH_i \odot \sqrt{ \dfrac{[\bs\Psi^{\top} \boldX]^{\text{pos}} + [\bs\Psi^{\top} \bs\Psi]^{\text{neg}} \boldH_i}{[\bs\Psi^{\top} \boldX]^{\text{neg}} + [\bs\Psi^{\top} \bs\Psi]^{\text{pos}} \boldH_i} }.
  \label{eq:H_update_rule}
\end{equation}

\subsection*{Complexity} % (fold)
\label{sub:complexity}

The computational complexity for the pre-training stage of Deep \seminmf is of order $\mathcal{O}\left( m t \left( pnk + nk^2 + kp^2 + kn^2 \right) \right)$, where $m$ is the number of
layers, $t$ the number of iterations until convergence and $k$ is the maximum number of components out of all the layers. The complexity for the fine-tuning stage is $\mathcal{O}\left(m t_f \left(
pnk + (p + n)k^2 \right)\right)$ where $t_f$ is the number of additional iterations needed.

\footnotetext{
The implementation and documentation of \autoref{algo:deepnmfalgo} can be found at \mbox{\tt\url{http://trigeorgis.com/deepseminmf}}.}

\subsection{Non-linear Representations}\label{sub:nonlinear}

By having a linear decomposition of the initial data distribution we may fail to describe efficiently the non-linearities that exist in between the latent attributes of the model. Introducing non-linear functions between the layers, can enable us to extract features for each of the latent attributes of the model that are non-linearly separable in the initial input space. 

This is motivated further from neurophysiology paradigms, as the theoretical and experimental evidence suggests that the human visual system has a hierarchical  and rather non-linear approach \cite{Riesenhuber1999} in processing image structure, in which neurons become selective to process progressively more complex features of the image structure. As argued by Malo {\it et al.} \cite{Malo2006}, employing an adaptive non-linear image representation algorithm results in a reduction of the statistical and the perceptual redundancy amongst the representation elements. 

From a mathematical point of view, one can use a non-linear function $g(\cdot)$, between each of the implicit representations ($\boldH_1^+, \ldots, \boldH_{m-1}^+$), in order to better approximate the non-linear manifolds which the given data matrix $\boldX$ originally lies on. In other words by using a non-linear squashing function we enhance the expressibility of our model and allow for a better reconstruction of the initial data. This has been proved in \cite{hornik1989multilayer} by the use of the Stone-Weierstrass theorem, in the case of multilayer feedforward network structures, which Semi-NMF is an instance of, that arbitrary squashing functions can approximate virtually any function of interest to any desired degree of accuracy, provided sufficiently many hidden units are available. 

To introduce non-linearities in our model we modify the $i^{\text{th}}$ feature matrix ${\bs H}_i$, by setting
\begin{equation} 
{\bs H}_{i} \approx g(\boldZ_{i+1} \boldH_{i+1}).
\end{equation}
which in turns changes the objective function of the model  to be:
\begin{equation}
C^* = \frac{1}{2} \norm{\boldX - \boldZ_1 g\left( \boldZ_2\; g\left( \cdots g\left( \boldZ_m \boldH_m \right)\right)\right)}^2_F
\end{equation}

In order to compute the derivative for the $i_{th}$ feature layer, we make use of the chain rule and get:
\begin{align*}
\dfrac{\partial\bs C^*}{\partial \boldH_i} &= \bs Z^{\top}_i \dfrac{\partial\bs C^*}{\partial {\boldZ_i \boldH_i}}\\
 &= \bs Z^{\top}_i \left[\dfrac{\partial\bs C^*}{\partial g\left({\boldZ_i \boldH_i}\right)} \odot \nabla g\left({\boldZ_i \boldH_i}\right)\right]\\
 &= \bs Z^{\top}_i \left[\dfrac{\partial\bs C^*}{\partial \bs H_{i-1}} \odot \nabla g\left({\boldZ_i \boldH_i}\right)\right]
\end{align*}
The derivation of the first feature layer $\bs H_1$ is then identical to the version of the model with one layer. 
\begin{align*}
\dfrac{\partial\bs C^*}{\partial \boldH_1} &= \frac{1}{2} \dfrac{\partial  Tr[-2 \bs X^{\top}\boldZ_1 \bs H_1 + (\boldZ_1 \bs H_1)^{\top} \boldZ_1 \bs H_1]}{\partial \boldH_1} \\
& = \bs Z_1^{\top} \bs Z_1 \bs {H}_1 - \bs Z_1^{\top} \bs X\\
& = \bs Z_1^{\top} \left(\bs Z_1 \bs {H}_1  - \bs X \right).
\end{align*}
Similarly we can compute the derivative for the weight matrices $\bs Z_i$, 
\begin{align*}
\dfrac{\partial\bs C^*}{\partial \boldZ_i} &= \dfrac{\partial\bs C^*}{\partial {\boldZ_i \bs{ H}_i}} \bs{ H}^{\top}_i \\
 &= \left[\dfrac{\partial\bs C^*}{\partial g\left({\boldZ_i \bs{ H}_i}\right)} \odot \nabla g\left({\boldZ_i \bs{ H}_i}\right)\right] \bs{ H}^{\top}_i \\
 &= \left[\dfrac{\partial\bs C^*}{\partial \bs H_{i-1}} \odot \nabla g\left({\boldZ_i \bs{ H}_i}\right)\right] \bs{ H}^{\top}_i
\end{align*}
and
\begin{align*}
\dfrac{\partial\bs C^*}{\partial \boldZ_1} &= \frac{1}{2} \dfrac{\partial  Tr[-2 \bs X^{\top}\boldZ_1 \bs{\tilde{H}}_1 + (\boldZ_1 \bs{\tilde{H}}_1)^{\top} \boldZ_1 \bs{\tilde H}_1]}{\partial \boldZ_1} \\
& =   \left(\bs Z_1 \bs{\tilde{H}}_1 - \bs X\right) \bs{\tilde{H}}_1^\top
\end{align*}

Using these derivatives we can make use of gradient descent optimizations such as Nesterov's optimal gradient \cite{nesterov2007gradient}, to minimize the cost function with respect to each of the weights of our model.

\section{Weakly-Supervised Attribute Learning}\label{sec:apriori}

As before, consider a dataset of faces $\bs X$ as in \autoref{fig:intuition}. In this dataset, we have a collection of subjects, where each one has a number of images expressing different expressions, taken by different angles (pose information). A three layer \deepseminmf model could be used here to automatically learn representations in an unsupervised manner ($\bs H_{\text{pose}}, \bs H_{\text{expression}}, \bs H_{\text{identity}}$) that conform to this latent hierarchy of attributes. Of course, the features are extracted without accounting (partially) available information that may exist for each of the these attributes of the dataset.

\begin{figure*}[hptb]
\centering
\includegraphics[width=\linewidth]{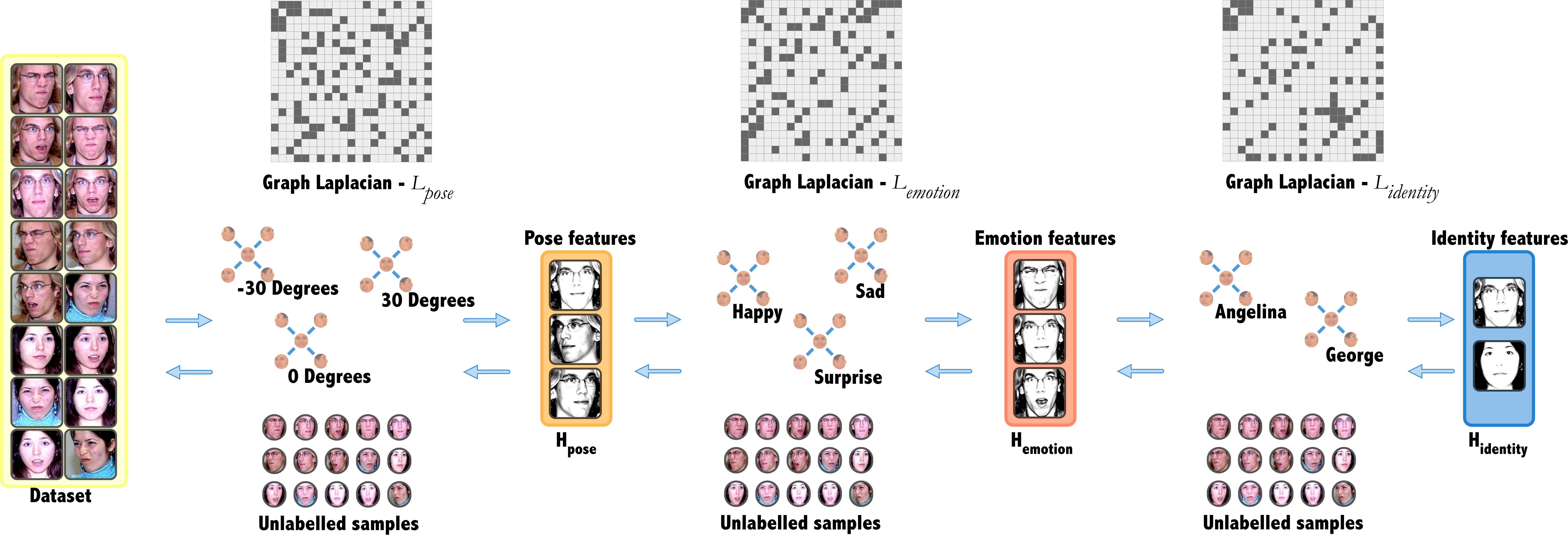}
\caption{A weakly-supervised Deep Semi-NMF model uses prior knowledge we have about the attributes of our model to improve the final representation of our data. In this illustration we incorporate information from pose, expression, and identity attributes into the 3 feature layers of our model $\bs H_{\text{pose}}$, $\bs H_{\text{expression}}$, and $\bs H_{\text{identity}}$ respectively. }
\label{fig:attribute_enrichment}
\end{figure*}

To this effect we propose a Deep Semi-NMF approach that can incorporate partial attribute information that we named \textsl{Weakly-Supervised Deep Semi-Nonnegative Matrix Factorization} (\dwsf). \dwsf is able to learn, in a semi-supervised manner, a hierarchy of representations; each level of this hierarchy corresponding to a specific attribute for which we may have only partial labels for. As depicted in \autoref{fig:attribute_enrichment}, we show that by incorporating some label information via graph regularization techniques we are able to do better than the \deepseminmf for classifying faces according to pose, expression, and identity. We also show that by initializing a \deepseminmf with the weights learned by a \dwsf we are able to improve the performance of the \deepseminmf for the task of clustering faces according to identity.

\subsection{Incorporating known attribute information}

Consider that we have an undirected graph $\bs G$ with $N$ nodes, where each of the nodes corresponds to one data point in our initial dataset. A node $i$ is connected to another node $j$ iff we have a priori knowledge that those samples share the same label, and this edge has a weight $\bs w_{ij}$.

In the simplest case scenario, we use a binary weight matrix $\bs W$ defined as:

\begin{equation}\label{eq:construct_w}
  \bs W_{ij} = \left\{ 
  \begin{array}{l l}
    1 & \quad \text{if $y_i$ = $y_j$}\\
    0 & \quad \text{otherwise}
  \end{array} \right. 
\end{equation}

Instead one can also choose a {\it radial basis function kernel} \begin{equation}\label{eq:construct_w_radial}
  \bs W_{ij} = \left\{ 
  \begin{array}{l l}
     \exp\left(-\frac{\norm{\bs{x}_i - \bs{x}_j}^2}{2\sigma^2}\right) & \quad \text{if $y_i$ = $y_j$}\\
    0 & \quad \text{otherwise}
  \end{array} \right. 
\end{equation} or a {\it dot-product weighting}, where 
\begin{equation}\label{eq:construct_w_un}
  \bs W_{ij} = \left\{ 
  \begin{array}{l l}
     \bs{x}_i^{\top} \bs{x}_j & \quad \text{if $y_i$ = $y_j$}\\
    0 & \quad \text{otherwise}
  \end{array} \right. 
\end{equation}

Using the graph weight matrix $\bs W$, we formulate $\bs{L}$, which denotes the Graph Laplacian \cite{cvetkovic1980spectra} that stores our prior knowledge about the relationship of our samples and is defined as $\bs{L} = \bs{D} - \bs{W}$, where $\bs{D}$ is a diagonal matrix whose entries are column (or row, since $\bs W$ is symmetric) sums of $\bs{W}$, $\bs{D}_{jj} =  \sum_k \bs{W}_{jk}$. In order to control the amount of embedded information in the graph we introduce as in \cite{Belkin2002,Belkin2006,Hao2013}, a term $\mathcal R$ which controls the smoothness of the low dimensional representation. \begin{align} \notag
  \mathcal{R} &= \sum_{j,l=1}^{N} \norm{\bs h_j - \bs h_l}^2 \bs W_{jl} \\\notag
  &= \sum_{j=1}^N \bs h_j^{\top} \bs h_j \bs D_{jj}-  \sum_{j,l=1}^{N} \bs h_j^{\top} \bs h_j \bs W_{jl} \\\notag
  &= \tr\left(\bs H^{\top} \bs D \bs H\right) - \tr\left(\bs H^{\top} \bs W \bs H\right)  \\ \label{eq:term_r}
  &= \tr\left(\bs H^{\top} \bs L \bs H\right)
\end{align}
where $h_i$ is the low-dimensional features for sample $i$, that we obtain from the decomposed model.

Minimizing this term $\mathcal R$, we ensure that the euclidean difference between the final level representations of any two data points $\bs h_i$ and $\bs h_j$ is low when we have prior knowledge that those samples have a relationship, producing similar features $\bs h_i$  and $\bs h_j$. On the other hand, when we do not have any expert information about some or even all the class information about an attribute, the term has no influence on the rest of the optimization.

\label{sec:wsnmf}
Before deriving the update rules and the algorithm for the multi-layer \dwsf{} model, we first show the simpler case of the one layer version, which will come into use for pre-training the model, as Semi-NMF can be used to pre-train the purely unsupervised Deep Semi-NMF. We call this model {\it Weakly Supervised Semi-NMF} \wsf{}. 

By combining the term $\mathcal R$ introduced in \autoref{eq:term_r}, with the cost function of \seminmf we obtain the cost function for Weakly-Supervised Factorization (\wsf).

\begin{align}\notag
C_{\text{\wsf}}=\|{\bs X} - {\bs Z^{\pm}}{\bs H^{+}}\|^2_F + \lambda \tr(\bs H^\top \bs L \bs H)\\ \label{eq:obj_wsnmf} \text{s.t.}\; \bs H \geq 0.
\end{align}

% \begin{figure}[hptb]
% \centering
% \includegraphics[width=0.95\linewidth]{wsnmf}
% \caption{An illustration of Weakly Supervised Semi-NMF model performs a decomposition of a face dataset, using a small amount of labeled samples which are incorporated in a Graph Laplacian ($\bs L$) to produce a final representation $\bs H$ better suited for identity classification.}
% \label{fig:wsnmf}
% \end{figure}
 
The update rules, but also the algorithm for training a \wsf model can be found in the supplementary material.

% \wsf{} uses the cost function of Semi-NMF while weakly enforcing the label constrains through the use of graph regularization techniques in a manner similar to GNMF \cite{gnmf}. This allows the model to incorporate prior infromation for a known attribute of a dataset, in a semi-supervised way. 

% . Instead of incorporating prior knowledge, using hard constrains such as in the case of CNMF \cite{Liu2012}, it weakly enforces the label constrains through the use of graph regularization techniques similar to GNMF \cite{gnmf}. CNMF uses the label information as hard constraints to incorporate the information, which we believe restricts the abilities of the model for representing out-of-sample data points.

We incorporate the available partial labelled information for the pose, expression, and identity by forming a graph Laplacian for pose for the first layer $(\bs L_{\text{pose}})$, expression for the second layer $(\bs L_{\text{expression}})$, and identity for the third layer $(\bs L_{\text{identity}})$ of the model. We can then tune the regularization parameters $\lambda_i$ accordingly for each of the layers to express the importance of each of these parameters to the \dwsf{} model. Using the modified version of our objective function \autoref{eq:apriori_cost}, we can derive the \autoref{fig:apriori_deepnmfalgo}.

\begin{align} \label{eq:apriori_cost} \notag
C_{\text{\dwsf}} = \frac{1}{2}\|{\bs X} - \bs Z_1 g\left(\ldots g\right(\bs Z_m {\bs H_m\left)\right)}\|^2_F  \\ + \frac{1}{2}\sum_{i=1}^{m} \lambda_i \tr(\bs H_i^\top \bs L_i \bs H_i)
\end{align}
In order to compute the derivative for the $i_{th}$ feature layer, we make use of the chain rule and get:
\begin{align*}
    \dfrac{\partial\Cwdeep}{\partial\bs H_i} &= \bs Z^{\top}_i \dfrac{\partial\Cdeep}{\partial {\bs Z_i \bs H_i}} + \frac{1}{2}\dfrac{\lambda_i \tr(\bs H_i^\top \bs L_i \bs H_i)}{\partial \bs H_i}  \\
 &= \bs Z^{\top}_i \left[\dfrac{\partial\Cdeep}{\partial \bs H_{i-1}} \odot \nabla g\left({\bs Z_i \bs H_i}\right)\right] \\ & \qquad +   \lambda_i \bs L_i \bs H_i
\end{align*}
And the derivation of the first feature layer $\bs H_1$ is then:
\begin{align*}
\dfrac{\partial\Cwdeep}{\partial \bs H_1} &=  \dfrac{\partial\Cdeep}{\partial {\bs Z_i \bs H_i}} + \frac{1}{2} \dfrac{\lambda_1 \tr(\bs H_1^\top \bs L_1 \bs H_1)}{\partial \bs H_1} \\
& = \bs Z_1^{\top} \left(\bs Z_1 \bs {H}_1  - \bs X \right) + \lambda_1 \bs L_1 \bs H_1.
\end{align*}
Similarly we can compute the derivative for the weight matrices $\bs Z_i$, 
\begin{align*}
\dfrac{\partial\Cwdeep}{\partial \bs Z_i} &= \dfrac{\partial\Cdeep}{\partial {\bs Z_i \bs{ H}_i}} \bs{ H}^{\top}_i \\
 &= \left[\dfrac{\partial\Cdeep}{\partial \bs H_{i-1}} \odot \nabla g\left({\bs Z_i \bs{ H}_i}\right)\right] \bs{ H}^{\top}_i
\end{align*} and \begin{align*}
\dfrac{\partial\Cwdeep}{\partial \bs Z_1} &= \dfrac{\partial \Cdeep}{\partial \bs Z_1} \\
& =   \left(\bs Z_1 \bs{\tilde{H}}_1 - \bs X\right) \bs{\tilde{H}}_1^\top 
\end{align*}
Using these derivatives we can make use of gradient descent optimizations as with the non-linear Deep Semi-NMF model, to minimize the cost function with respect to each of the factors of our model. If instead use the linear version of the algorithm where $g$ is the identity function, then we can derive a multiplicative update algorithm version of \dwsf, as described in \autoref{fig:apriori_deepnmfalgo}.

\begin{algorithm}[tb]
\caption{Proposed algorithm for training a \dwsf~model. Initially we approximate the factors greedily using WSF or Semi-NMF and we fine-tune the factors until we reach the convergence criterion.}

\begin{algorithmic}
    \label{fig:apriori_deepnmfalgo}
    % \FUNCTION {\textsc{DeepSemiNMF}}
        \STATE {\bfseries Input:} {${\boldsymbol X} \in \mathbb{R}^{p\times n}$, list of layer sizes {\it layers}}
        \STATE {\bfseries Output:} {weight matrices ${\boldsymbol Z}_i$ and feature matrices ${\boldsymbol H}_i$ for each of the layers}
        \STATE {}
        \STATE {Initialize Layers}
        \FORALL{layers}
            \STATE ${\boldsymbol Z}_i, {\boldsymbol H}_i \gets  $ {\sc WSF}(${\boldsymbol H}_{i-1}$, layers($i$), $\lambda_i$) 
        \ENDFOR

        \STATE{}

        \REPEAT
            \FORALL{layers}
                \STATE
       $\tilde{{\boldsymbol H}_i} \gets \left\{
          \begin{array}{l l}
            {\boldsymbol H}_i           & \quad \text{\textbf{if}} \quad i = k \\
            {\boldsymbol Z}_{i+1} \tilde{\boldsymbol H}_{i+1} & \quad \text{otherwise}\\
          \end{array}\right.$ \\
       ${\boldsymbol \Psi} \gets  \prod_{k=1}^{i-1}{{\boldsymbol Z}_k}$ \\
       ${\boldsymbol Z}_i \gets {\boldsymbol \Psi}^{\dagger} {\boldsymbol X} \tilde{\boldsymbol H}_i^{\dagger}$ \\

       \newcommand{\hNum}{[{\boldsymbol \Psi}^{\top} {\boldsymbol X}]^{\text{pos}} + [{\boldsymbol \Psi}^{\top} {\boldsymbol \Psi}]^{\text{neg}} {\boldsymbol H}_i + \lambda_i \bs{H}_i \bs{W}_{i}}
       \newcommand{\hDenum}{[{\boldsymbol \Psi}^{\top} {\boldsymbol X}]^{\text{neg}} + [{\boldsymbol \Psi}^{\top} {\boldsymbol \Psi}]^{\text{pos}} {\boldsymbol H}_i + \lambda_i \bs{H}_i \bs{D}_i}
       
       $\bs{F} \gets \dfrac{\hNum}{\hDenum}$ \\
       
        ${\boldsymbol H}_i \gets  {\boldsymbol H}_i \odot \bs{F}^{\eta}$
            \ENDFOR
        \UNTIL{Stopping criterion is reached}
    % \ENDFUNCTION
    \end{algorithmic}
\end{algorithm}

\subsection{Weakly Supervised Factorization with Multiple Label Constraints}
Another approach we propose within this framework is a single--layer \wsf model that learns only a single representation based on information from multiple attributes. This Multiple-Attribute extension of the \wsf, the  \mawsf, accounts for the case of having multiple number of attributes $\xi$ for our data matrix $\bs X$, by having a regularization term $\lambda_i \tr[\bs H \bs L_i \bs H^{\top}]$. This term uses the prior information from all the available attributes to construct $\xi$ Laplacian graphs where each of them has a different regularization factor $\lambda_i$.

This constitutes \mawsf, whose cost function is
\begin{align}\notag
C_{\text{mawsf}}=\|{\bs X} - {\bs Z}{\bs H}\|^2_F + \sum_{i=1}^\xi \lambda_i &\tr(\bs H^{\top} \bs L_i \bs H)\\ &\text{s.t.}\; \bs H \geq 0
\end{align}

The update rules used, and the algorithm can be found in the supplementary material.

\section{Out-of-sample projection} % (fold)
\label{sub:project_out_of_sample_data}

After learning an internal model of the data, either using the purely unsupervised \deepseminmf or to perform semi-supervised learning using the \dwsf~ model  with learned weights $\bs Z$, and features $\bs H$ we can project an out-of-sample data point $\bs x^{\ast}$ to the new lower-dimensional embedding $\bs h^{\ast}$. We can accomplish this using one of the two presented methods.

\vspace{0.5cm}

\noindent {\bfseries\scshape Method 1: Basis matrix reconstruction.}\\ Each testing sample $\bs x^{\ast}$ is projected into the linear space defined by the weights matrix $\bs Z$. Although this method has been used by various previous works \cite{Turk1991a,Li2001} using the NMF model, it does {\it not} guarantee the non-negativity of $\bs h^{\ast}$.

For the linear case of \dwsf, this would lead to

\begin{equation}
  \bs h^{\ast} \approx \left[\bs Z_1 \bs Z_2 \ldots \bs Z_l\right]^{\dagger} \bs x^{\ast}.
\end{equation}

and for the non-linear case

\begin{equation}
  \bs h^{\ast} \approx g^{-1} \left(\bs Z_l^{\dagger} \left( \cdots \left(\bs Z_2^\dagger g^{-1}\left(\bs Z_1^\dagger \bs x^{\ast}\right)\right)\right)\right)
\end{equation}

\def\hnew{\bs h^{\ast}}

\noindent {\bfseries\scshape Method 2: Using non-negativity update rules.}

Using the same process as in \deepseminmf, we can intuitively learn the new features $\hnew$, by assuming that the weight matrices $\forall i. \bs Z_i$ remain fixed.

% {\bfseries linear case:}
\begin{align}\label{eq:project_deep}
 \forall l. \bs h_l^{\ast} = \argmin_{\bs h} \norm{\bs x^{\ast} - \prod_{i=1}^l \bs Z_i \bs h_l} \\\notag \text{such that} \; \bs h_l \geq 0.
\end{align}
and for the non-linear case
% {\bfseries non-linear case:}
\begin{align}\label{eq:project_deep_nonlinear}
 \forall l. \bs h_l^{\ast} = \argmin_{\bs h} \norm{\bs x^{\ast} - \bs Z_1 g\left(\bs Z_2 \cdots g\left(\bs Z_l \bs h_l\right)\right)} \\\notag \text{such that} \; \bs h_l \geq 0.
\end{align}

where $\bs h_l$, corresponds to the $l_{th}$ feature layer for the out-of-sample data point $\bs x^\ast$.  This problem is then solved by using \autoref{algo:deepnmfalgo} as Deep Semi-NMF, but without updating the weight matrices $\bs Z_i$.

\section{Experiments}\label{sec:experiments}
Our main hypothesis is that a \deepseminmf is able to learn better high-level representations of our original data than a one-layer \seminmf for clustering according to the attribute with the lowest variability in the dataset. In order to evaluate this hypothesis, we have compared the performance of Deep \seminmf with that of other methods, on the task of clustering images of faces in two distinct datasets. These datasets are:
\begin{itemize}
\item \textbf{CMU PIE: } We used a freely available version of CMU Pie ~\cite{PIEpami}, which comprises of $2,856$ grayscale $32\times32$ face images of 68 subjects. Each person has 42 facial
images under different light and illumination conditions. In this database we only know the identity of the face in each image.
\item \textbf{XM2VTS: } \textit{The Extended Multi Modal Verification for Teleservices and Security applications} (XM2VTS) \cite{messer1999xm2vtsdb} contains $2,360$ frontal images of $295$ different subjects. Each subject has two available images for each of the four different laboratory sessions, for a total of 8 images. The images were eye-aligned and resized to $42 \times 30$.
\end{itemize}

In order to evaluate the performance of our \deepseminmf model, we compared it against not only \seminmf \cite{ding2010convex}, but also against other NMF variants that could be useful in learning such representations.
More specifically, for each of our two datasets we performed the following experiments:
\begin{itemize}
\item \textbf{Pixel Intensities: } By using only the pixel intensities of the images in each of our datasets, which of course give us a strictly non-negative input data matrix $\boldX$, we compare the reconstruction error and the clustering performance of our Deep \seminmf method against the \seminmf, NMF with multiplicative update rules \cite{seung2001algorithms}, Multi-Layer NMF \cite{song2013hierarchical}, GNMF \cite{cai2011graph}, and NeNMF \cite{guan2012nenmf}.
\item \textbf{Image Gradient Orientations (IGO): } In general, the trend in Computer Vision is to use complicated engineered features like HoGs, SIFT, LBPs, etc. As a proof of concept, we choose to
conduct experiments with simple gradient orientations~\cite{tzimiropoulos2012subspace} as features, instead of pixel intensities, which results into a data matrix of mixed signs, and expect that we can learn better data
representations for clustering faces according to identities. In this case, we only compared our Deep \seminmf with its one-layer \seminmf equivalent, as the other techniques are not able to deal with
mixed-sign matrices.
\end{itemize}

In \autoref{sub:pretraining}, having demonstrated the effectiveness of the purely unsupervised \deepseminmf model we show next how pretraining a \dwsf~ model on an auxiliary dataset and using the learned weights to perform unsupervised \deepseminmf can lead to significant improvements in terms of the clustering accuracy. 

Finally, in \autoref{sub:multiple_attributes},  we examine the classification abilities of the proposed models for each of the three attributes of the CMU Multi-PIE dataset (pose/expression/identity) and use this to test more on our secondary hypothesis, i.e. that every representation in each layer is in fact most suited for learning according to the attributes that corresponds to the layer of interest.

\subsection{An example with multi-modal synthetic data}\label{sub:xor_problem}

\begin{figure}[h]
\centering
\def\svgwidth{.75\linewidth}
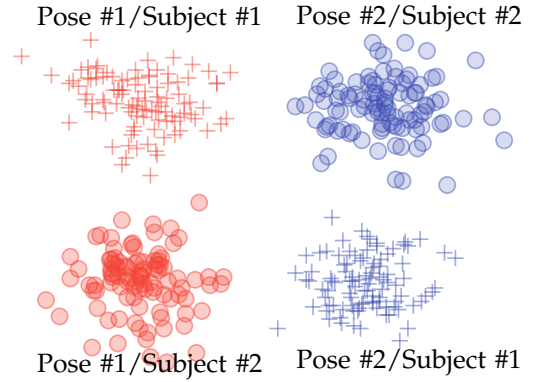
\caption{Visualisation of $\bs X_{\text{XOR}}$, where \tikzcircle{black} (\tikzplus{black}) are samples of Subject \#1 (Subject \#2) and red (blue) data points denote the samples of each subject with Pose \#1 (Pose \#2).}
\label{fig:xor_dataset}
\end{figure}

As previously mentioned images of faces are multi-modal distributions which are composed of multiple attributes such as pose and identity. A simplified example of such dataset is \autoref{fig:xor_dataset} where we have two subjects depicting two poses each. This example two-dimensional dataset $\bs X_{\text{XOR}}$  was generated using 100 samples from four normal distributions with $\sigma=1$. 

As previously discussed in \autoref{sub:nonlinear}, Semi-NMF is an instance of a single layer neural network. As such there can not exist a linear projection $\bs Z_{\text{XOR}}$ which maps the original data distribution $\bs X_{\text{XOR}}$ into a sub-space such as the two subjects (red and blue) of the dataset are linearly separable. 

\begin{figure}[h]
\centering
\subfloat[Layer \#1 features $\bs H_1$]{
\includegraphics[width=.49\linewidth]{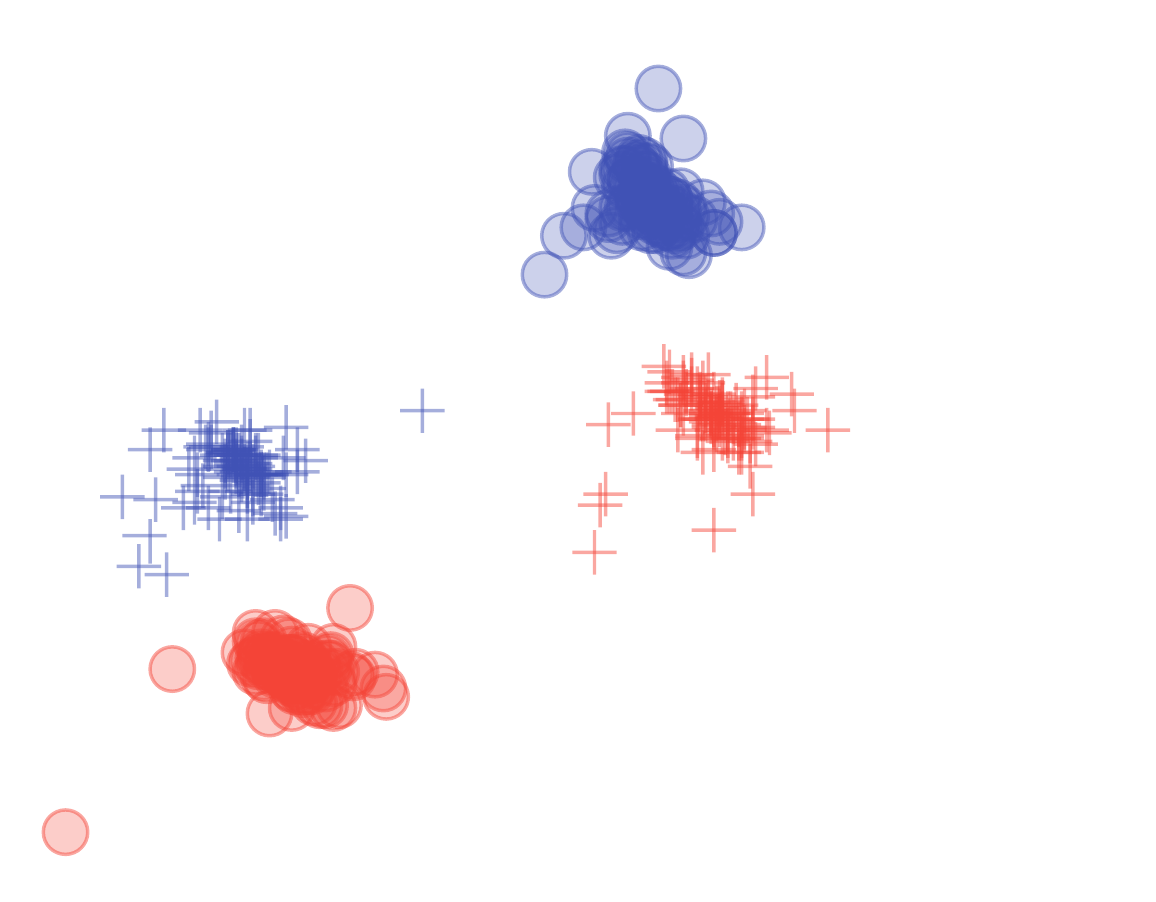}
}
\subfloat[Layer \#2 features $\bs H_2$]{
\includegraphics[width=.49\linewidth]{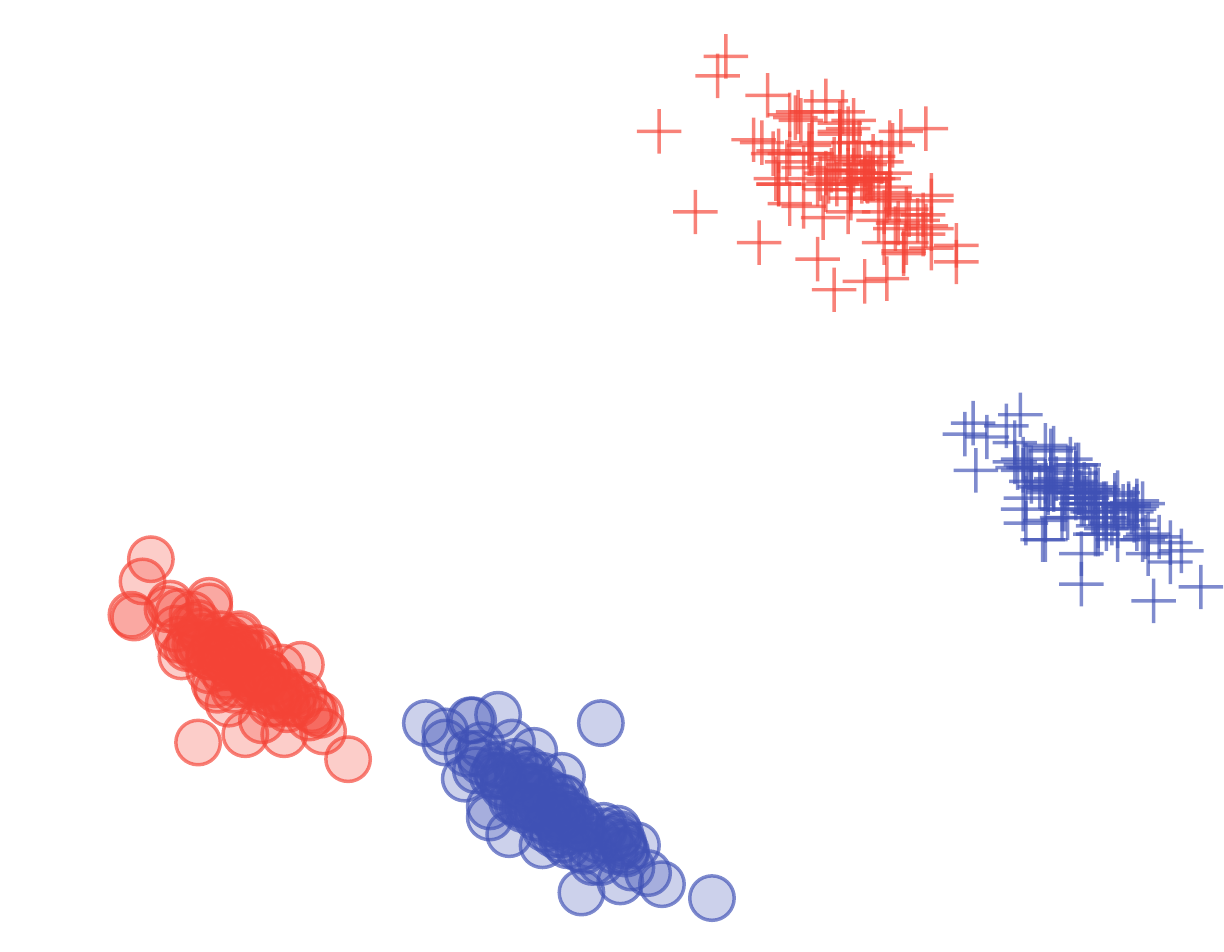}
}
\caption{The features extracted by each of the layers of a deep factorization model on the artificially generated dataset. The second layer manages to find a projection of the initial data which makes all the classes linearly separable, a task which is infeasible with a simple Semi-NMF model.} \label{fig:xor_features}
\end{figure}

Instead by employing a deep factorization model using the labels for the pose and identity for the first and second layer respectively we can find a non-linear mapping which separates the two identities as shown in \autoref{fig:xor_features}.

\vspace{-0.3cm}

\subsection{Implementation Details}
To initiate the matrix factorization process, NMF and \seminmf algorithms start from some initial point ($\boldZ^0, \boldH^0$), where usually $\boldZ^0$ and $\boldH^0$ are randomly initialized matrices. 

A problem with this approach, is not only the initialization point is far from the final convergence point, but also makes the process non deterministic.

The proposed initialization of \seminmf by its authors is instead by using the \emph{k}-means algorithm \cite{Ding2010}. Nonetheless, $k$-means is computationally heavy when the number of components $k$ is fairly high ($k>100$). As an alternative we implemented the approach by \cite{gillis2014exact} which suggests exact and heuristic algorithms which solve \seminmf decompositions using an SVD based initialization. We have found that using this method for Semi-NMF, Deep Semi-NMF, and \wsf{} helps the algorithms to converge a lot sooner.

Similarly, to speed up the convergence rate of NMF we use the Non-negative Double Singular Value decomposition (NNDSVD) suggested by Boutsidis \etal \citet{boutsidis2008svd}. NNDSVD is a method based on
two SVD processes, one to approximate the initial data matrix $\boldX$ and the other to approximate the positive sections of the resulting partial SVD factors. 

For the GNMF experimental setup, we chose a suitable number of neighbours to create the regularizing graph, by visualizing our datasets using Laplacian Eigenmaps~\cite{belkin2001laplacian}, such that we had visually distinct clusters (in our case 5).

\subsection{Number of layers}

\begin{figure}[hpt]
\centering
\subfloat[XM2VTS]{
            \includegraphics[trim=2mm 1mm 1mm 0,clip,width=.49\linewidth]{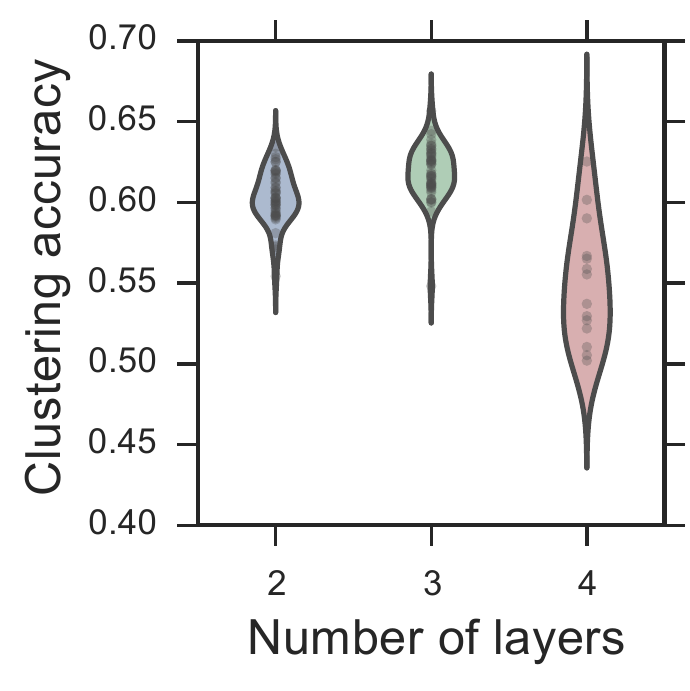}
            \label{fig:xm2vts_alllayers}
}
\subfloat[CMU PIE]{
      \includegraphics[trim=2mm 1mm 1mm 0,clip,width=.49\linewidth]{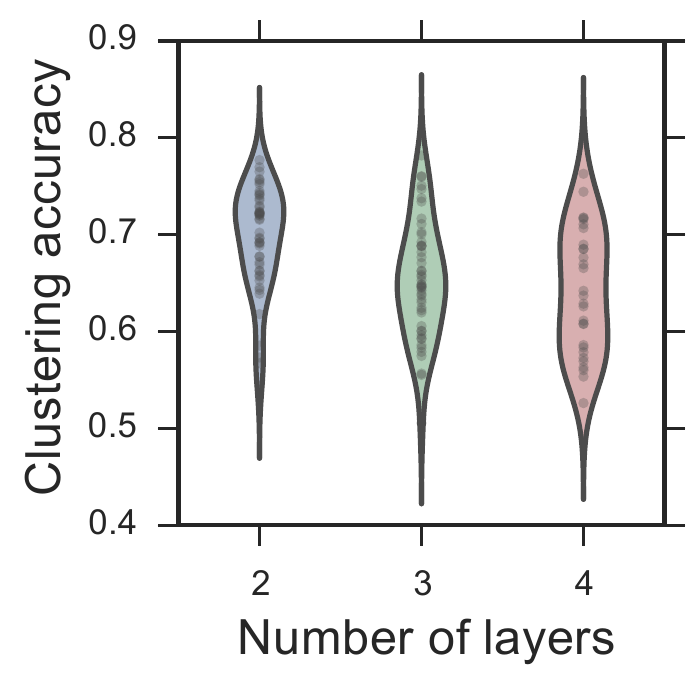}
            \label{fig:multipie_alllayers}
}
\caption{{\bfseries Number of layers \emph{vs.} clustering accuracy}
We generated two hundred configurations of the \deepseminmf with a variant number of layers, and we evaluate the the final feature layer $\bs H_m$ according to its clustering accuracy for the XM2VTS (left) and CMU PIE (right) datasets. To make these models comparable we keep a constant number of components for the last layer($40$), and we generated the number of components for the rest of the layers according to an exponential distribution with a mean of 400 components.}
\label{fig:all_layers}
\end{figure}

Important for the experimental setup is the selected structure of the multi-layered models. After careful preliminary experimentation, we focused on experiments that involve two hidden layer architectures for the Deep \seminmf and Multi-layer NMF. We specifically experimented with models that had a first hidden representation $\boldH_1$ with $625$ features, and a second representation $\boldH_2$ with a number of features that ranged from $20$ to $70$. This allowed us to have comparable configurations between the different datasets and it was a reasonable compromise between speed and accuracy. Nonetheless, in \autoref{fig:all_layers} we show experiments with more than two layers on our two datasets. In the latter experiment we generated two hundred configurations of the \deepseminmf with a variant number of layers, and we evaluated the the final feature layer $\bs H_m$ according to its clustering accuracy for the XM2VTS and \cmupie datasets. To make these models comparable we keep a constant number of components for the last layer ($40$) and we generated the number of components for the rest of the layers drawn from an exponential distribution with a mean of 400 components and then arrange them in an decreasing order. We decided to do so to comply with our main assumption: the first layers of our hierarchical model capture attributes with a larger variance and thus the model needs a larger capacity to encode them, where as the last layers will capture attributes with a lower variance. 

\begin{figure}[tb]
     \centering
     \includegraphics[width=\linewidth,clip]{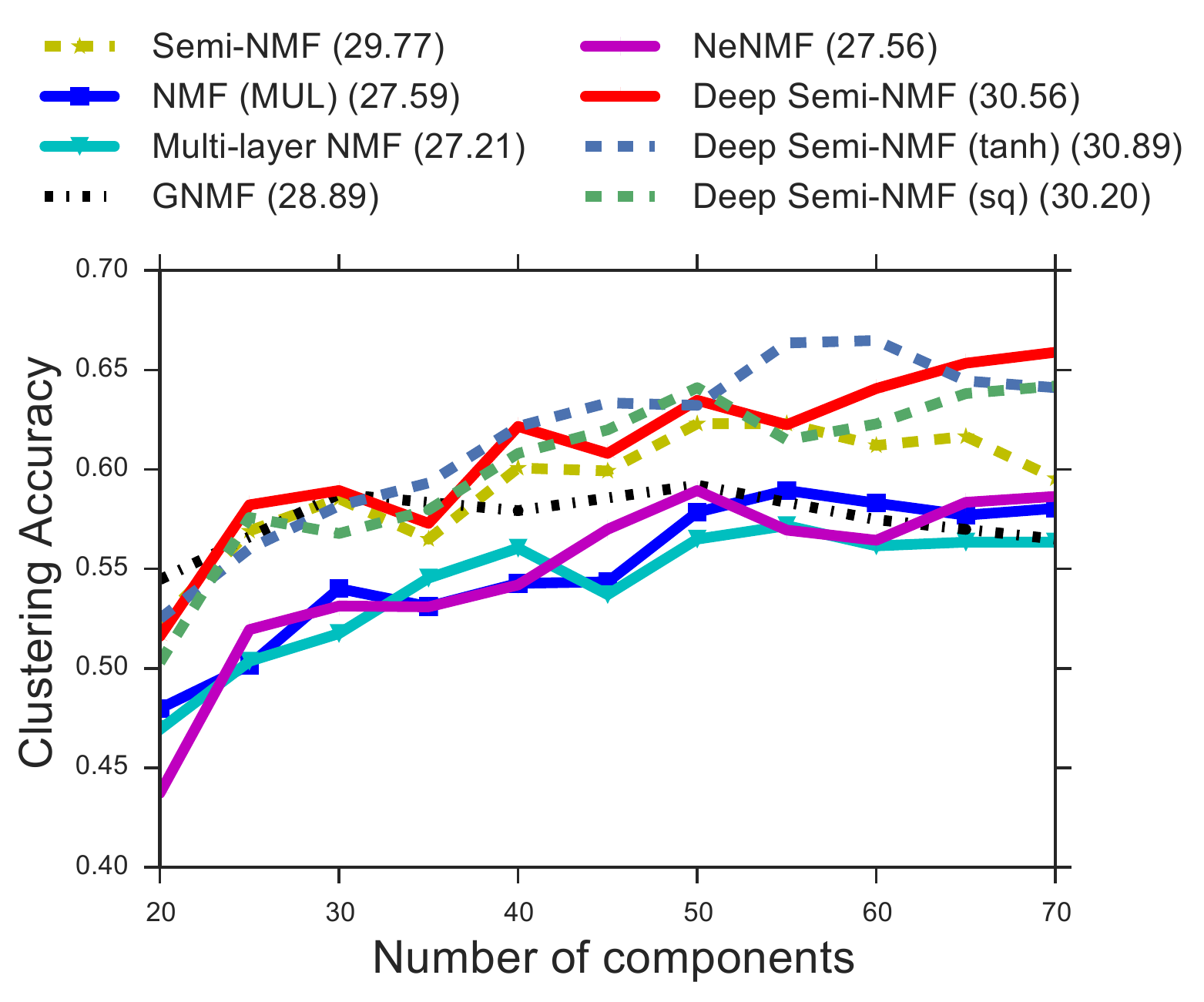}
     \vspace{-2em}
     \caption{\textbf{XM2VTS-Pixel Intensities:} Accuracy for clustering based on the representations learned by each model with respect to identities. The deep architectures are comprised of 2 representation layers (1260-625-$a$) and the representations used were from the top layer. In parenthesis we show the AUC scores.}
     \label{fig:xm2vts_intensities}
\end{figure}
\begin{figure}[ht]
     \centering
     \includegraphics[width=\linewidth,clip]{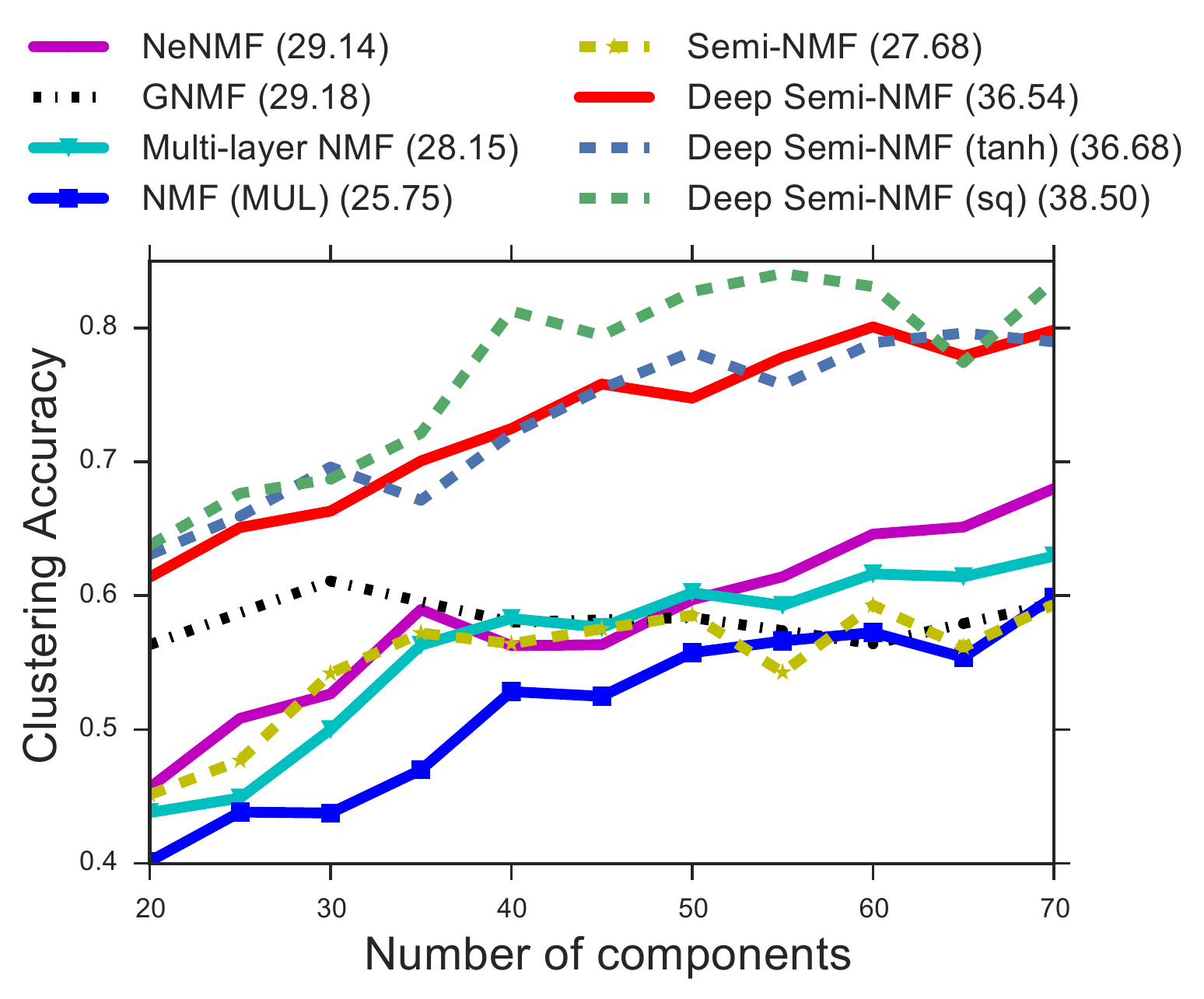}
     \vspace{-2em}
     \caption{\textbf{CMU PIE--Pixel Intensities: } Accuracy for clustering based on the representations learned by each model with respect to identities. The deep architectures are comprised of 2 representation layers (1024-625-$a$) and the representations used were from the top layer. In parenthesis we show the AUC scores.}
     \label{fig:pie_intensities}
\end{figure}

\subsection{Reconstruction Error Results}
Our first experiment was to evaluate whether the extra layers, which naturally introduce more factors and are therefore more difficult to optimize, result in a lower quality local optimum. We evaluated how well the matrix decomposition is performed by calculating the reconstruction error, the Frobenius norm of the difference between the original data and the reconstruction for all the
methods we compared. Note that, in order to have comparable results, all of the methods have the same stopping criterion rules. We have set the maximum amount of iterations to 1000 (usually ${\sim}100$
iterations are enough) and we use the convergence rule $E_{i-1} - E_i \leq \kappa \text{ max}(1, E_{i-1})$ in order to stop the process when the reconstruction error $(E_i)$ between the current and previous update is small enough. In our experiments we set $\kappa = 10^{-6}$. \autoref{tab:reconstruction_error} shows the change in reconstruction error with respect to the selected number of features in $\boldH_2$ for all the methods we used on the Multi-PIE dataset. 

\begin{table}[hptb]\label{tab:reconstruction_error}
\renewcommand{\arraystretch}{1.1}
\small
\begin{tabular}{l|rrrrrr}
 & \multicolumn{6}{c}{\# Components} \\ \cline{2-7}
 \bfseries Model &  \bfseries  20 &  \bfseries   30 &  \bfseries  40 &  \bfseries  50 &  \bfseries  60 & \bfseries   70 \\
\midrule\midrule
 Deep Semi-NMF   &   9.18 &   7.61 &  6.50 &  5.67 &  4.99 &  4.39 \\
 GNMF            &  10.56 &   9.35 &  8.73 &  8.18 &  7.81 &  7.48 \\
 Multi-layer NMF &  11.11 &  10.16 &  9.28 &  8.49 &  7.63 &  6.98 \\
 NMF (MUL)       &  10.53 &   9.36 &  8.51 &  7.91 &  7.42 &  7.00 \\
 NeNMF           &   9.83 &   8.39 &  7.39 &  6.60 &  5.94 &  5.36 \\
 Semi-NMF        &   9.14 &   7.57 &  6.43 &  5.53 &  4.76 &  4.13 \\
 \midrule
\bottomrule
\end{tabular}
\caption{The reconstruction error ($\norm{\bs X - \tilde{\bs X}}^2_F$) for each of the algorithms on the CMU PIE dataset, for a variable number of components.}
\end{table}

The results show that \seminmf manages to reach a much lower reconstruction error than the other methods consistently, which would match our expectations as it does not constrain the weights $\boldZ$ to be non-negative. What is important to note here is that the Deep \seminmf models do not have a significantly lower reconstruction error compared to the equivalent \seminmf models, even though the approximation involves more factors. Multi-layer NMF and GNMF have a larger reconstruction error, in return for uncovering more meaningful features than their NMF counterpart.

\subsection{Clustering Results}
After achieving satisfactory reconstruction error for our method, we proceeded to evaluate the features learned at the final representation layer, by using $k$-means clustering, as in \cite{cai2011graph}. To assess the clustering quality of the representations produced by each of the algorithms we compared, we take advantage of the fact that the datasets are already labelled. The two metrics used were the accuracy (AC) and the normalized mutual information metric (NMI), as those are defined in \cite{xu2003document}. For a cleaner presentation we have included all the experiments that use NMI in the supplement.

We made use of two main non-linearities for our experiments, the scaled hyperbolic tangent $stanh(x) = \alpha \text{tanh}(\beta x)$ with $\alpha = 1.7159, \beta = \frac{2}{3}$ \cite{LeCun2012}, and a square auxiliary function $sq(x) = x^2$. 

Figures \ref*{fig:xm2vts_intensities}-\ref*{fig:pie_intensities} show the comparison in clustering accuracy when using $k$-means on the feature representations produced by each of the techniques we compared, when our input matrix contained only the pixel intensities of each image. Our method significantly outperforms every method we compared it with on all the datasets, in terms of clustering accuracy.
 
\begin{figure}
     \centering
     \includegraphics[width=\linewidth,clip]{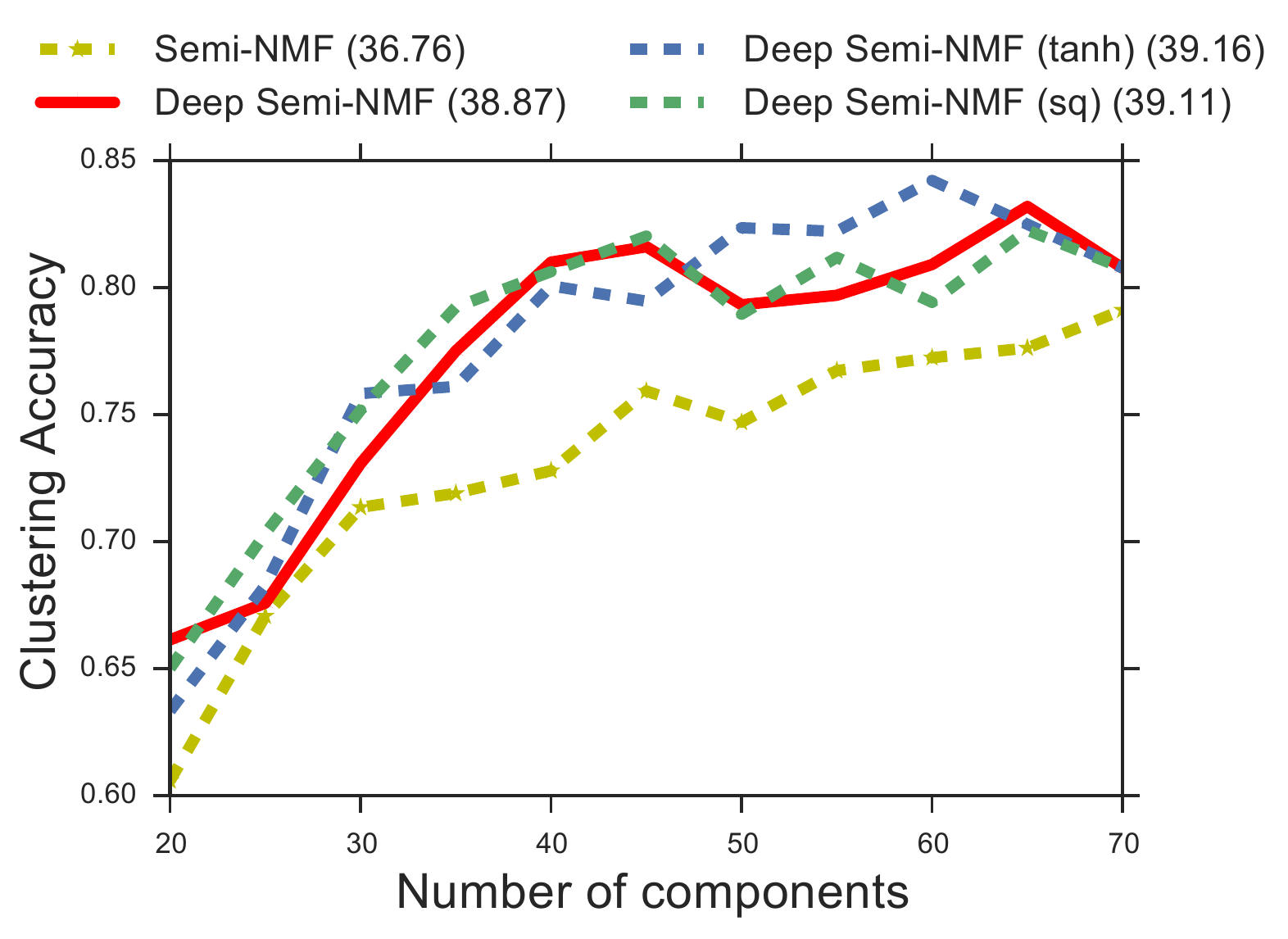}
     \vspace{-2em}
     \caption{\textbf{XM2VTS-IGO: } Accuracy scores on clustering based on the representations learned by each model with respect to identities. The deep architectures are comprised of 2 hidden layers (2520-625-$a$) and the representations used were from the top layer. In parenthesis we show the AUC scores.}
     \label{fig:xmvts_igo}
\end{figure}

\begin{figure}[tb]
     \centering
     \includegraphics[width=\linewidth,clip]{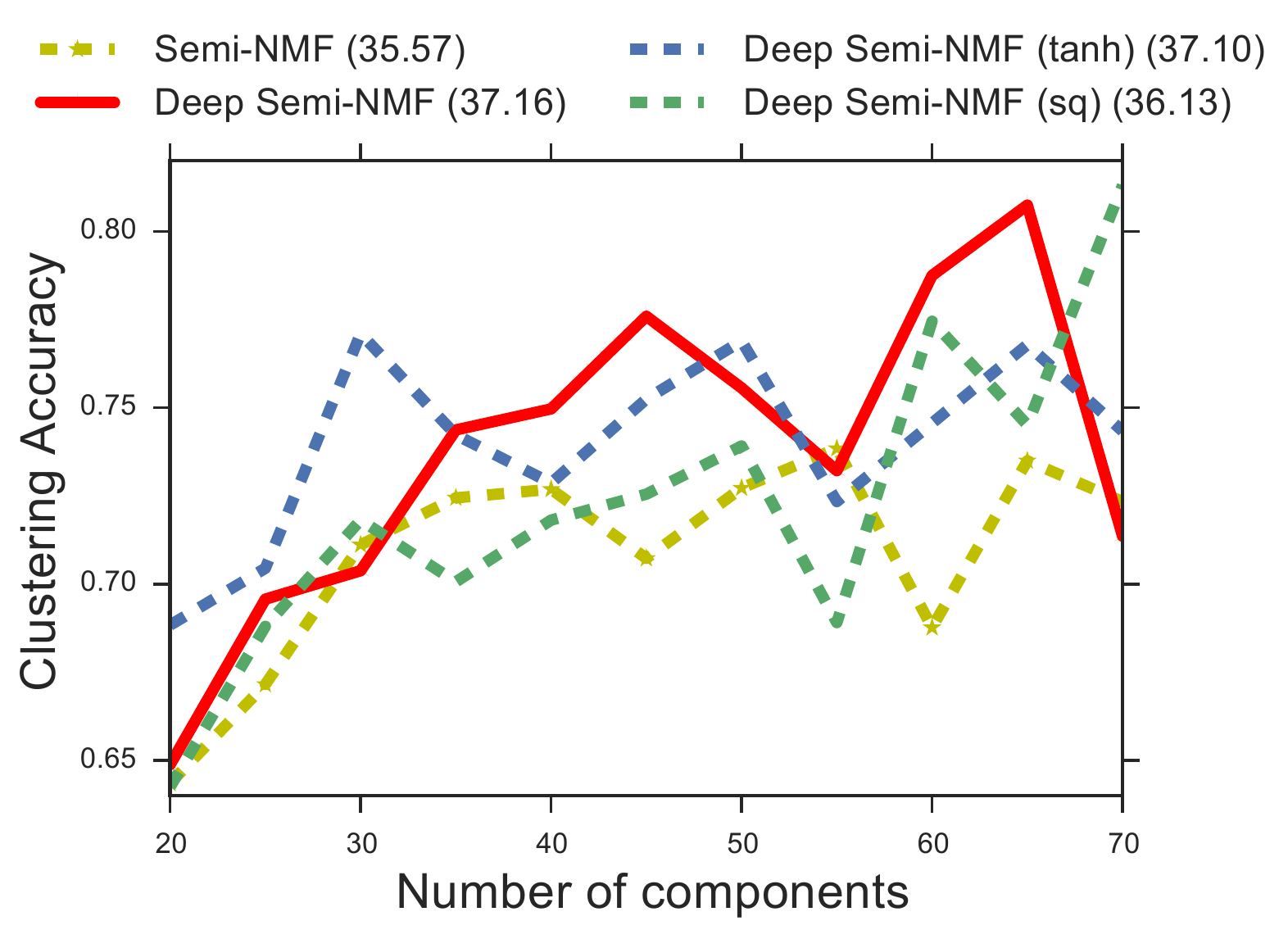}
    \vspace{-2em}
    \caption{\textbf{CMU PIE-IGO: } Accuracy scores on clustering based on the representations learned by each model with respect to identities. The deep architectures are comprised of 2 hidden layers (2048-625-$a$) and the representations used were from the top layer. In parenthesis we show the AUC scores.}
     \label{fig:pie_igo}
\end{figure}

By using IGOs, the Deep \seminmf was able to outperform the single-layer \seminmf as shown in Figures \ref*{fig:xmvts_igo}-\ref*{fig:pie_igo}. Making use of these simple mixed-signed features improved the clustering accuracy considerably. It should be noted that in all cases, with the exception of the
CMU PIE Pose experiment with IGOs, our Deep \seminmf outperformed all other methods with a difference in performance that is statistically significant (paired t-test, $p
\ll 0.01$).


\subsection{Supervised pre-training}\label{sub:pretraining}

As the optimization process of deep architectures is highly non-convex, the initialization point of the process is an important factor for obtaining good final representation for the initial dataset. Following trends in deep learning \cite{bengio2007greedy}, we show that supervised pretraining of our model on a auxiliary dataset and using the learned weights as initialization points for the unsupervised \deepseminmf algorithm can lead to significant performance improvements in regards to clustering accuracy.

As an auxiliary dataset we use XM2VTS where we resize all the images to a 32x32 resolution to match the CMU PIE image resolution, which is our primary dataset.
Splitting the XM2VTS dataset to training/validation sets, we learn weights $\bs Z^{\text{xm2vts}}_{1,2}$ using a \dwsf{} model with (625--$a$) layers, and regularization parameters $\lambda = \{0, 0.01\}$.

We then use the obtained weights $\bs Z^{\text{xm2vts}}_{1,2}$ from the supervised task as an initialization point and perform unsupervised fine-tuning on the CMU PIE dataset. To evaluate the resulting features, we once again perform clustering using the $k$-means algorithm.

\begin{figure}[hptb]
  \centering
  \includegraphics[width=\linewidth]{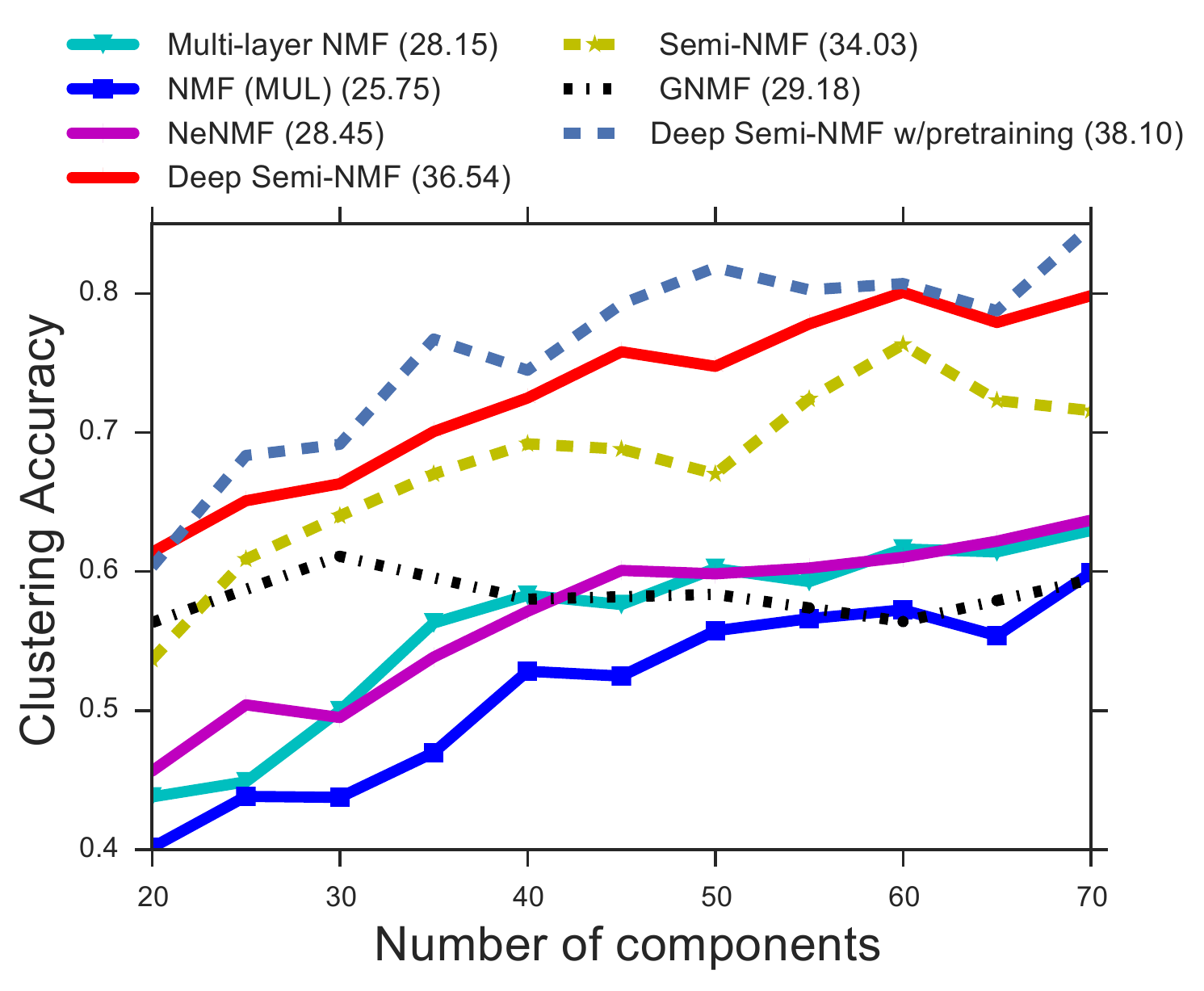}
  \vspace{-2em}
  \caption{\textbf{Supervised pre-training:} Clustering accuracy on the CMU PIE dataset, after supervised training on the XM2VTS dataset using a priori \deepseminmf. In parenthesis we show the AUC scores.} \label{fig:pretraining}
\end{figure}

In our experiments all the models  with supervised pre-training outperformed the ones without, as shown in \autoref{fig:pretraining}, in terms of clustering accuracy. Additionally this validates our claim of how pretraining can be exploited to get better representations out of unlabelled data.

\subsection{Learning with Respect to Different Attributes}\label{sub:multiple_attributes}
Finally, we conducted experiments for classification using each of the three representations learned by our three-layered \dwsf models when the input was the raw pixel intensities of our images of a larger subset of the CMU Multi-PIE dataset.

CMU Multi-PIE contains around $750,000$ images of $337$ subjects, captured under laboratory conditions in four different sessions. In this work, we used a subset of $7,905$ images of $147$ subjects in $5$ different poses  and expressing 6 different emotions, which is the amount of samples that we had annotations and were imposed to the same illumination conditions. Using the annotations from  \citet{Sagonas2013,Sagonas2013a}, we aligned these images based on a common frame. After that, we resized them to a smaller resolution of $40\times 30$. The database comes with labels for each of the attributes mentioned above: identity, illumination, pose, expression.  We only used CMU Multi-PIE for this experiment since we only had identity labels for our other datasets. We split this subset into a training and validation set of 2025 images, and the rest for testing.

\begin{table}[!Ht]
\renewcommand{\arraystretch}{2.2}
\centering
\begin{tabular}{r|l|c||c||c}
\hline
& \bfseries  Model & \bfseries Pose & \bfseries Expression & \bfseries Identity\\
\hline\hline
\multirow{3}{*}{\begin{sideways}\small Unsupervised\end{sideways}} & Semi-NMF    &  99.73  & 81.50 & 36.46 \\
&NMF         & 100.00 & 80.68 & 49.12 \\
&\deepseminmf &  99.86 & 80.54 & 61.22 \\
\midrule
\multirow{2}{*}{\begin{sideways}\small Semi\end{sideways}} &CNMF        & 89.21 & 33.88 & 28.30 \\
&DNMF        & 100.00 & 82.22 & 55.78 \\
\midrule
\multirow{3}{*}{\begin{sideways}\small Proposed \end{sideways}}  &\wsf      & 100.00 & 81.50 & 63.81 \\
&\mawsf & 100.00  & 81.50 & 64.08 \\
&\dwsf   & \bfseries 100.00 & \bfseries 82.90 & \bfseries 65.17  \\
\hline
\end{tabular} 
\caption{The performance in accuracy on the CMU Multi-PIE dataset using an SVM classifier on top of the features learned. For the multi-layer models we used 3 layers corresponding to each of the attributes, and performed classification using the features learned for the corresponding attribute. For the one-layer models, we learned three different representations, one for each layer.}
\label{tab:three_layer}
\end{table}

We compare the classification performance of an SVM classifier (with a penalty parameter $\gamma = 1$) using the data representations of the NMF, Semi-NMF, and \deepseminmf models that have no attribute information. The CNMF \cite{Liu2012}, DNMF \cite{Kotsia2007}, and our WSF models that have attribute labels only for the attribute we were classifying for,  and our \mawsf{} and \dwsf{} that learned data representations based on all attribute information available. In \autoref{tab:three_layer}, we demonstrate the performance in accuracy of each of the methods. In all of the methods, each feature layer has 100 components, and in the case of the \dwsf{} model, we have used $\forall i . \lambda_i = 10^{-3}$. 

We also compared the performance of our \dwsf with that of \wsf and \mawsf to see whether the different levels of representation amount to better performance in classification tasks for each of the attributes represented.  In both cases, but also in comparison with the rest state-of-the-art unsupervised and semi-supervised matrix factorization techniques, our proposed solution manages to extract better features for the task at hand as seen in \autoref{tab:three_layer} for classification.
\begin{figure}[hptb]
  \centering
  \includegraphics[width=0.95\linewidth]{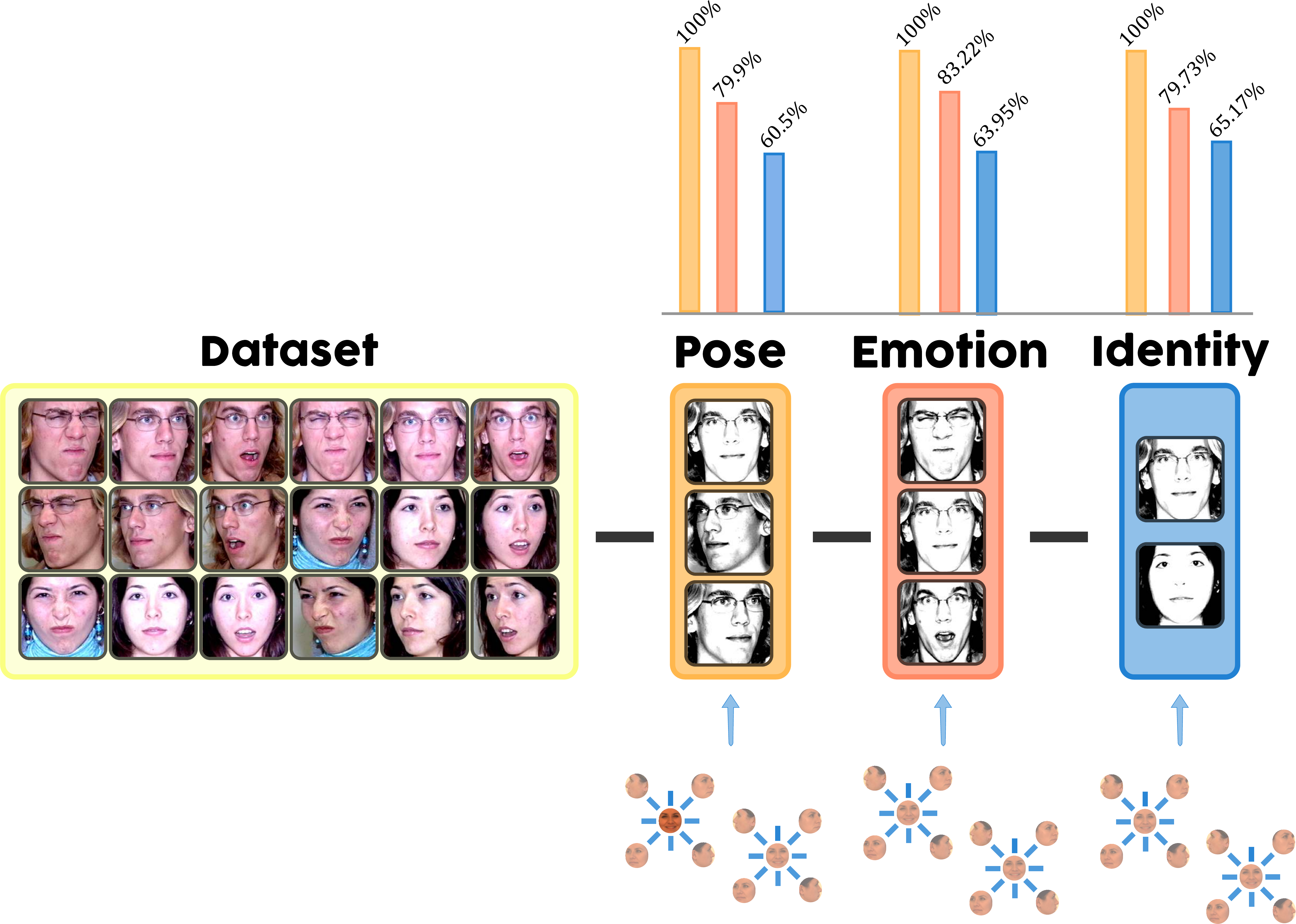}
  \caption{A three layer \dwsf~model trained on \cmupie~with only frontal illumination (camera 5). The bars depict the accuracy levels for the pose~(\tikzcircle{pose}), emotion~(\tikzcircle{emotion}), and identity~(\tikzcircle{identity}) respectively, for each layer, with a linear SVM classifier.}
  \label{fig:apriori_results}
\end{figure}
\section{Conclusion} % (fold)
\label{sec:conclusion}
We have introduced a novel deep architecture for semi-non-negative matrix factorization, the Deep Semi-NMF, that is able to automatically learn a hierarchy of attributes of a given dataset, as well as
representations suited for clustering according to these attributes. Furthermore we have presented an algorithm for optimizing the factors of our Deep \seminmf, and we evaluate its performance compared to
the single-layered \seminmf and other related work, on the problem of clustering faces with respect to their identities. We have shown that our technique is able to learn a high-level, final-layer
representation for clustering with respect to the attribute with the lowest variability in the case of two popular datasets of face images, outperforming the considered range of typical powerful NMF-based techniques.

We further proposed \dwsf{}, which incorporates knowledge from the known attributes of a dataset that might be available. \dwsf can be used for datasets that have (partially) annotated attributes or even are a combination of different data sources with each one providing different attribute information. We have demonstrated the abilities of this model on the CMU Multi-PIE dataset, where using additional information provided to us during training about the pose, emotion, and identity information of the subject we were able to uncover better features for each of the attributes, by having the model learning from all the available attributes simultaneously. Moreover, we have shown that \dwsf{} could be used to pretrain models on auxiliary datasets, not only to speed up the learning process, but also uncover better representations for the attribute of interest.

Future avenues include experimenting with other applications, e.g. in the area of speech recognition, especially for multi-source speech recognition and we will investigate multilinear extensions of the proposed framework \cite{zafeiriou2009discriminant,zafeiriou2009algorithms}.

% use section* for acknowledgement
\ifCLASSOPTIONcompsoc
  % The Computer Society usually uses the plural form
  \section*{Acknowledgments}
\else
  % regular IEEE prefers the singular form
  \section*{Acknowledgment}
\fi

\label{sec:acknowledgements}
George Trigeorgis is a recipient of the fellowship of the Department of Computing, Imperial College London, and this work was partially funded by it. The work of Konstantinos Bousmalis was funded partially from the Google Europe Fellowship in Social Signal Processing. The work of Stefanos Zafeiriou was partially funded by the EPSRC project EP/J017787/1 (4D-FAB). The work of Bj\"orn~W.~Schuller was partially funded by the European Community's Horizon 2020 Framework Programme under grant agreement No. 645378 (ARIA-VALUSPA). The responsibility lies with
the authors.

% Can use something like this to put references on a page
% by themselves when using endfloat and the captionsoff option.
\ifCLASSOPTIONcaptionsoff
  \newpage
\fi

% trigger a \newpage just before the given reference
% number - used to balance the columns on the last page
% adjust value as needed - may need to be readjusted if
% the document is modified later
%\IEEEtriggeratref{8}
% The "triggered" command can be changed if desired:
%\IEEEtriggercmd{\enlargethispage{-5in}}

% references section

\bibliographystyle{IEEEtran}
\bibliography{tpami}

% Generated by IEEEtran.bst, version: 1.13 (2008/09/30)
\begin{thebibliography}{10}
\providecommand{\url}[1]{#1}
\csname url@samestyle\endcsname
\providecommand{\newblock}{\relax}
\providecommand{\bibinfo}[2]{#2}
\providecommand{\BIBentrySTDinterwordspacing}{\spaceskip=0pt\relax}
\providecommand{\BIBentryALTinterwordstretchfactor}{4}
\providecommand{\BIBentryALTinterwordspacing}{\spaceskip=\fontdimen2\font plus
\BIBentryALTinterwordstretchfactor\fontdimen3\font minus
  \fontdimen4\font\relax}
\providecommand{\BIBforeignlanguage}[2]{{%
\expandafter\ifx\csname l@#1\endcsname\relax
\typeout{** WARNING: IEEEtran.bst: No hyphenation pattern has been}%
\typeout{** loaded for the language `#1'. Using the pattern for}%
\typeout{** the default language instead.}%
\else
\language=\csname l@#1\endcsname
\fi
#2}}
\providecommand{\BIBdecl}{\relax}
\BIBdecl

\bibitem{paatero1994positive}
P.~Paatero and U.~Tapper, ``Positive matrix factorization: A non-negative
  factor model with optimal utilization of error estimates of data values,''
  \emph{Environmetrics}, vol.~5, no.~2, pp. 111--126, 1994.

\bibitem{brunet2004metagenes}
J.-P. Brunet, P.~Tamayo, T.~R. Golub, and J.~P. Mesirov, ``Metagenes and
  molecular pattern discovery using matrix factorization,'' \emph{PNAS}, vol.
  101, no.~12, pp. 4164--4169, 2004.

\bibitem{devarajan2008nonnegative}
K.~Devarajan, ``Nonnegative matrix factorization: an analytical and
  interpretive tool in computational biology,'' \emph{PLoS computational
  biology}, vol.~4, no.~7, p. e1000029, 2008.

\bibitem{berry2005email}
M.~W. Berry and M.~Browne, ``Email surveillance using non-negative matrix
  factorization,'' \emph{Computational \& Mathematical Organization Theory},
  vol.~11, no.~3, pp. 249--264, 2005.

\bibitem{zafeiriou2006exploiting}
S.~Zafeiriou, A.~Tefas, I.~Buciu, and I.~Pitas, ``Exploiting discriminant
  information in nonnegative matrix factorization with application to frontal
  face verification,'' \emph{TNN}, vol.~17, no.~3, pp. 683--695, 2006.

\bibitem{kotsia2007novel}
I.~Kotsia, S.~Zafeiriou, and I.~Pitas, ``A novel discriminant non-negative
  matrix factorization algorithm with applications to facial image
  characterization problems.'' \emph{TIFS}, vol.~2, no. 3-2, pp. 588--595,
  2007.

\bibitem{weninger2012optimization}
F.~Weninger and B.~Schuller, ``{Optimization and parallelization of monaural
  source separation algorithms in the openBliSSART toolkit},'' \emph{Journal of
  Signal Processing Systems}, vol.~69, no.~3, pp. 267--277, 2012.

\bibitem{ding2010convex}
C.~H. Ding, T.~Li, and M.~I. Jordan, ``Convex and semi-nonnegative matrix
  factorizations,'' \emph{IEEE TPAMI}, vol.~32, no.~1, pp. 45--55, 2010.

\bibitem{ding2005equivalency}
C.~Cing, X.~He, and H.~Simon, ``On the equivalence of nonnegative matrix
  factorization and spectral clustering,'' in \emph{Proc. {SIAM} Data Mining},
  2005.

\bibitem{Herrero2001}
J.~Herrero, A.~Valencia, and J.~Dopazo, ``{A hierarchical unsupervised growing
  neural network for clustering gene expression patterns.}''
  \emph{Bioinformatics (Oxford, England)}, vol.~17, pp. 126--136, 2001.

\bibitem{Zhao2005}
Y.~Zhao and G.~Karypis, ``{Hierarchical clustering algorithms for document
  datasets},'' \emph{Data Mining and Knowledge Discovery}, vol.~10, pp.
  141--168, 2005.

\bibitem{Tsoumakas2007}
G.~Tsoumakas and I.~Katakis, ``{Multi-label classification: An overview},''
  \emph{International Journal of Data Warehousing and Mining}, vol.~3, pp.
  1--13, 2007.

\bibitem{Zhang2010}
Y.~Zhang and Z.-H. Zhou, ``{Multilabel dimensionality reduction via dependence
  maximization},'' pp. 1--21, 2010.

\bibitem{Trigeorgis2014}
G.~Trigeorgis, K.~Bousmalis, S.~Zafeiriou, and W.~B. Schuller, ``{A Deep
  Semi-NMF Model for Learning Hidden Representations},'' in \emph{ICML},
  vol.~32, 2014.

\bibitem{hinton2006reducing}
G.~E. Hinton and R.~R. Salakhutdinov, ``Reducing the dimensionality of data
  with neural networks,'' \emph{Science}, vol. 313, no. 5786, pp. 504--507,
  2006.

\bibitem{golub1970singular}
G.~H. Golub and C.~Reinsch, ``Singular value decomposition and least squares
  solutions,'' \emph{Numerische Mathematik}, vol.~14, no.~5, pp. 403--420,
  1970.

\bibitem{wold1987principal}
S.~Wold, K.~Esbensen, and P.~Geladi, ``Principal component analysis,''
  \emph{Chemometrics and intelligent laboratory systems}, vol.~2, no.~1, pp.
  37--52, 1987.

\bibitem{seung2001algorithms}
D.~D. Lee and H.~S. Seung, ``Algorithms for non-negative matrix
  factorization,'' \emph{Advances in neural information processing systems},
  vol.~13, pp. 556--562, 2001.

\bibitem{cai2011graph}
D.~Cai, X.~He, J.~Han, and T.~S. Huang, ``Graph regularized nonnegative matrix
  factorization for data representation,'' \emph{TPAMI}, vol.~33, no.~8, pp.
  1548--1560, 2011.

\bibitem{ahn2004multiplicative}
J.-H. Ahn, S.~Choi, and J.-H. Oh, ``A multiplicative up-propagation
  algorithm,'' in \emph{ICML}.\hskip 1em plus 0.5em minus 0.4em\relax ACM,
  2004, p.~3.

\bibitem{lyu2013algorithms}
S.~Lyu and X.~Wang, ``On algorithms for sparse multi-factor nmf,'' in
  \emph{Advances in Neural Information Processing Systems}, 2013, pp. 602--610.

\bibitem{Cichocki06multilayernonnegative}
A.~Cichocki and R.~Zdunek, ``Multilayer nonnegative matrix factorization,''
  \emph{Electronics Letters}, vol.~42, pp. 947--948, 2006.

\bibitem{song2013hierarchical}
H.~A. Song and S.-Y. Lee, ``Hierarchical data representation model -
  multi-layer nmf,'' \emph{ICLR}, vol. abs/1301.6316, 2013.

\bibitem{Liu2012}
H.~Liu, Z.~Wu, X.~Li, D.~Cai, and T.~S. Huang, ``{Constrained Nonnegative
  Matrix Factorization for Image Representation},'' \emph{PAMI}, vol.~34, pp.
  1299--1311, 2012.

\bibitem{Kotsia2007}
I.~Kotsia, S.~Zafeiriou, and I.~Pitas, ``{Novel discriminant non-negative
  matrix factorization algorithm with applications to facial image
  characterization problems},'' \emph{TIFS}, vol.~2, pp. 588--595, 2007.

\bibitem{Riesenhuber1999}
M.~Riesenhuber and T.~Poggio, ``{Hierarchical models of object recognition in
  cortex.}'' \emph{Nature neuroscience}, vol.~2, no.~11, pp. 1019--1025, 1999.

\bibitem{Malo2006}
J.~Malo, I.~Epifanio, R.~Navarro, and E.~P. Simoncelli, ``{Nonlinear image
  representation for efficient perceptual coding},'' \emph{TIP}, vol.~15, pp.
  68--80, 2006.

\bibitem{hornik1989multilayer}
K.~Hornik, M.~Stinchcombe, and H.~White, ``Multilayer feedforward networks are
  universal approximators,'' \emph{Neural networks}, vol.~2, no.~5, pp.
  359--366, 1989.

\bibitem{nesterov2007gradient}
Y.~Nesterov \emph{et~al.}, ``Gradient methods for minimizing composite
  objective function,'' 2007.

\bibitem{cvetkovic1980spectra}
D.~M. Cvetkovic, M.~Doob, and H.~Sachs, \emph{{Spectra of graphs: Theory and
  application}}.\hskip 1em plus 0.5em minus 0.4em\relax Academic press New
  York, 1980, vol. 413.

\bibitem{Belkin2002}
M.~Belkin and P.~Niyogi, ``{Using Manifold Structure for Partially Labelled
  Classification},'' in \emph{NIPS 2002}, 2002, pp. 271--277.

\bibitem{Belkin2006}
M.~Belkin, P.~Niyogi, and V.~Sindhwani, ``{Manifold regularization: A geometric
  framework for learning from labeled and unlabeled examples},'' \emph{JMLR},
  vol.~7, pp. 2399--2434, 2006.

\bibitem{Hao2013}
Y.~Hao, C.~Han, G.~Shao, and T.~Guo, ``{Generalized graph regularized
  non-negative matrix factorization for data representation},'' in
  \emph{Lecture Notes in Electrical Engineering}, vol. 210 LNEE, 2013, pp.
  1--12.

\bibitem{Turk1991a}
M.~Turk and A.~Pentland, ``{Eigenfaces for Recognition},'' 1991.

\bibitem{Li2001}
S.~Li, X.~W. H. X.~W. Hou, H.~J. Z. H.~J. Zhang, and Q.~S. C. Q.~S. Cheng,
  ``{Learning spatially localized, parts-based representation},'' \emph{CVPR},
  vol.~1, 2001.

\bibitem{PIEpami}
T.~Sim, S.~Baker, and M.~Bsat, ``"the cmu pose, illumination, and expression
  database.",'' \emph{TPAMI}, vol.~25, no.~12, pp. 1615--1618, 2003.

\bibitem{messer1999xm2vtsdb}
K.~Messer, J.~Matas, J.~Kittler, J.~Luettin, and G.~Maitre, ``Xm2vtsdb: The
  extended m2vts database,'' in \emph{International conference on audio and
  video-based biometric person authentication}, vol. 964.\hskip 1em plus 0.5em
  minus 0.4em\relax Citeseer, 1999, pp. 965--966.

\bibitem{guan2012nenmf}
N.~Guan, D.~Tao, Z.~Luo, and B.~Yuan, ``{NeNMF: an optimal gradient method for
  nonnegative matrix factorization},'' \emph{TSP}, vol.~60, no.~6, pp.
  2882--2898, 2012.

\bibitem{tzimiropoulos2012subspace}
G.~Tzimiropoulos, S.~Zafeiriou, and M.~Pantic, ``Subspace learning from image
  gradient orientations,'' 2012.

\bibitem{Ding2010}
C.~Ding, T.~Li, and M.~I. Jordan, ``{Convex and semi-nonnegative matrix
  factorizations},'' \emph{TPAMI}, vol.~32, pp. 45--55, 2010.

\bibitem{gillis2014exact}
N.~Gillis and A.~Kumar, ``{Exact and Heuristic Algorithms for Semi-Nonnegative
  Matrix Factorization},'' \emph{arXiv preprint arXiv:1410.7220}, 2014.

\bibitem{boutsidis2008svd}
C.~Boutsidis and E.~Gallopoulos, ``Svd based initialization: A head start for
  nonnegative matrix factorization,'' \emph{Pattern Recognition}, vol.~41,
  no.~4, pp. 1350--1362, 2008.

\bibitem{belkin2001laplacian}
M.~Belkin and P.~Niyogi, ``Laplacian eigenmaps and spectral techniques for
  embedding and clustering.'' in \emph{NIPS}, vol.~14, 2001, pp. 585--591.

\bibitem{xu2003document}
W.~Xu, X.~Liu, and Y.~Gong, ``Document clustering based on non-negative matrix
  factorization,'' in \emph{SIGIR}.\hskip 1em plus 0.5em minus 0.4em\relax ACM,
  2003, pp. 267--273.

\bibitem{LeCun2012}
Y.~A. LeCun, L.~Bottou, G.~B. Orr, and K.~R. M\"{u}ller, ``{Efficient
  backprop},'' \emph{Lecture Notes in Computer Science (including subseries
  Lecture Notes in Artificial Intelligence and Lecture Notes in
  Bioinformatics)}, vol. 7700 LECTU, pp. 9--48, 2012.

\bibitem{bengio2007greedy}
Y.~Bengio, P.~Lamblin, D.~Popovici, and H.~Larochelle, ``Greedy layer-wise
  training of deep networks,'' \emph{Advances in neural information processing
  systems}, vol.~19, p. 153, 2007.

\bibitem{Sagonas2013}
C.~Sagonas, G.~Tzimiropoulos, S.~Zafeiriou, and M.~Pantic, ``{300 faces
  in-the-wild challenge: The first facial landmark Localization Challenge},''
  in \emph{CVPR}, 2013, pp. 397--403.

\bibitem{Sagonas2013a}
------, ``{A semi-automatic methodology for facial landmark annotation},'' in
  \emph{CVPR-W}, 2013, pp. 896--903.

\bibitem{zafeiriou2009discriminant}
S.~Zafeiriou, ``Discriminant nonnegative tensor factorization algorithms,''
  \emph{TNN}, vol.~20, no.~2, pp. 217--235, 2009.

\bibitem{zafeiriou2009algorithms}
------, ``Algorithms for nonnegative tensor factorization,'' in \emph{Tensors
  in Image Processing and Computer Vision}.\hskip 1em plus 0.5em minus
  0.4em\relax Springer, 2009, pp. 105--124.

\end{thebibliography}
\vspace{-1.3cm}
\begin{IEEEbiographynophoto}{George Trigeorgis}
 is pursuing a Ph.D. degree from Imperial College London.
\end{IEEEbiographynophoto}
\vspace{-1.2cm}
\begin{IEEEbiographynophoto}{Konstantinos Bousmalis}
is a researcher working with Google Robotics, California.
\end{IEEEbiographynophoto}
\vspace{-1.2cm}
\begin{IEEEbiographynophoto}{Stefanos Zafeiriou}
is a Lecturer in the Department of Computing, Imperial College London.
\end{IEEEbiographynophoto}
\vspace{-1.2cm}
\begin{IEEEbiographynophoto}{Bj\"orn~W.~Schuller}
is a Lecturer in the Department of Computing, Imperial College London.
\end{IEEEbiographynophoto}

\vfill

% that's all folks
\end{document}